\documentclass[11pt]{article}
\usepackage{acl} 
\usepackage{times}
\usepackage{latexsym}
\usepackage[T1]{fontenc}
\usepackage[utf8]{inputenc}
\usepackage{microtype}
\usepackage{inconsolata}
\usepackage{graphicx}
\usepackage{booktabs}
\usepackage{multirow}
\usepackage{amsmath}
\usepackage{amsfonts}
\usepackage{xcolor}
\usepackage{tikz}
\usetikzlibrary{positioning}
\usepackage[most]{tcolorbox}
\usepackage[table]{xcolor}
\usepackage{caption}
\usepackage{placeins}
\usepackage{dblfloatfix} 
\usepackage{placeins}    
\usepackage{etoolbox}
\AtBeginEnvironment{table}{\small}
\usepackage{adjustbox}

\usepackage[most]{tcolorbox}
\usepackage{enumitem}

\usepackage[most]{tcolorbox}
\usepackage{xcolor}

\definecolor{casebg}{HTML}{F7F8FA}
\definecolor{caseborder}{HTML}{9AA4B2}
\definecolor{errorbg}{HTML}{FDECEC}
\definecolor{correctbg}{HTML}{EAF7EA}

\newtcolorbox{casebox}[1]{
  colback=casebg,
  colframe=caseborder,
  boxrule=0.5pt,
  arc=2pt,
  left=4pt,
  right=4pt,
  top=4pt,
  bottom=4pt,
  title=\textbf{#1},
  fonttitle=\small,
  coltitle=black,
  colbacktitle=casebg,
  breakable
}

\usepackage[most]{tcolorbox}
\usepackage{xcolor}

\definecolor{casebg}{HTML}{F8F9FA}
\definecolor{caseborder}{HTML}{9AA4B2}
\definecolor{workerbg}{HTML}{FFFFFF}
\definecolor{managerbg}{HTML}{F1F5F9}

\newtcolorbox{qualcase}[1]{
  colback=casebg,
  colframe=caseborder,
  boxrule=0.5pt,
  arc=2pt,
  left=5pt,
  right=5pt,
  top=5pt,
  bottom=5pt,
  title=\textbf{#1},
  fonttitle=\small,
  coltitle=black,
  colbacktitle=casebg,
  breakable
}

\newtcolorbox{workerbox}[1]{
  colback=workerbg,
  colframe=caseborder,
  boxrule=0.35pt,
  arc=1pt,
  left=4pt,
  right=4pt,
  top=3pt,
  bottom=3pt,
  title=\textbf{#1},
  fonttitle=\small,
  coltitle=black,
  colbacktitle=workerbg,
  breakable
}

\newtcolorbox{managerbox}{
  colback=managerbg,
  colframe=caseborder,
  boxrule=0.35pt,
  arc=1pt,
  left=4pt,
  right=4pt,
  top=3pt,
  bottom=3pt,
  breakable
}

\usepackage[most]{tcolorbox}
\usepackage{xcolor}

\definecolor{examplebg}{HTML}{F8F9FA}
\definecolor{exampleborder}{HTML}{9AA4B2}
\definecolor{rationalebg}{HTML}{FFFFFF}
\definecolor{takeawaybg}{HTML}{F1F5F9}

\newtcolorbox{planningexample}[1]{
  colback=examplebg,
  colframe=exampleborder,
  boxrule=0.5pt,
  arc=2pt,
  left=5pt,
  right=5pt,
  top=5pt,
  bottom=5pt,
  title=\textbf{#1},
  fonttitle=\small,
  coltitle=black,
  colbacktitle=examplebg,
  breakable
}

\newtcolorbox{rationalebox}{
  colback=rationalebg,
  colframe=exampleborder,
  boxrule=0.35pt,
  arc=1pt,
  left=4pt,
  right=4pt,
  top=3pt,
  bottom=3pt,
  breakable
}

\newtcolorbox{takeawaybox}{
  colback=takeawaybg,
  colframe=exampleborder,
  boxrule=0.35pt,
  arc=1pt,
  left=4pt,
  right=4pt,
  top=3pt,
  bottom=3pt,
  breakable
}

\newcommand{\labeltag}[2]{%
  \colorbox{#1}{\strut\footnotesize\textbf{#2}}%
}

\newtcolorbox{promptbox}[1]{%
  enhanced,
  breakable,
  colback=white,
  colframe=black!55,
  boxrule=0.45pt,
  arc=2pt,
  left=6pt,right=6pt,top=6pt,bottom=6pt,
  title=\textbf{#1},
  colbacktitle=black!5,
  coltitle=black,
  fonttitle=\normalsize,
  boxed title style={
    size=small,
    boxrule=0.45pt,
    colframe=black!55,
    colback=black!5,
    arc=2pt,
    left=6pt,right=6pt,top=2pt,bottom=2pt,
  },
  attach boxed title to top left={xshift=6pt,yshift=-1.5mm},
}
\setlist[itemize]{leftmargin=*,nosep}
\setlist[enumerate]{leftmargin=*,nosep}

\title{
Multi-Agent Reasoning with Adaptive Worker Allocation for Stance Detection              
}

\author{
Meysam Sabbaghan$^{1}$,
Arman Zareian Jahromi$^{1}$,
Doina Caragea$^{1}$ \\
$^{1}$Department of Computer Science \\
Kansas State University
}

\begin{document}
\maketitle

\begin{abstract}
Stance detection requires identifying an author's position toward a target, often from short-form texts where stance is implicit, indirect, or rhetorically framed. Although large language models (LLMs) achieve strong performance on this task, single-pass prompting can be brittle when multiple interpretations are plausible. Existing aggregation strategies, such as majority voting or self-consistency, improve robustness by combining labels, but they discard the intermediate reasoning needed to resolve conflicting interpretations.

We introduce a multi-agent reasoning framework with adaptive worker allocation for stance detection that shifts aggregation from label-level voting to reasoning-level synthesis. The framework employs a Manager--Worker architecture in which a Manager adaptively allocates a variable number of Worker agents based on input complexity. Each Worker analyzes the input from a distinct perspective and produces a reasoning-only explanation without emitting a stance label; the Manager then synthesizes these explanations to produce the final prediction.

We evaluate the proposed framework on SemEval-2016, P-Stance, and COVID-19 Stance using Llama, Mistral, and Gemini. Results show that the framework yields the largest gains on implicit and context-dependent stance cases, achieving 86.07 Macro-F1 on COVID-19 and 82.90 on SemEval-2016, while remaining competitive on more explicit stance datasets such as P-Stance. These findings suggest that adaptive reasoning-level aggregation is most beneficial when stance cannot be reliably inferred from surface cues alone.
\end{abstract}

\section{Introduction}

Stance detection aims to determine an author's position toward a specific
target, such as a public figure, policy, or ideological concept
\cite{mohammad2016semeval,sobhani2016detecting}. The task is especially
challenging in short-form online texts, where stance is often conveyed
implicitly through sarcasm, rhetorical framing, indirect references, or
attacks on related entities rather than through explicit statements
\cite{li2021p,glandt2021covidstance,bamman2015contextualized,riloff2013sarcasm}.

Unlike sentiment analysis, stance detection requires target-aware pragmatic
reasoning: the same expression may imply different stances depending on the
target. For example, criticizing a political opponent may indicate support
for the target, while an apparently positive statement may signal opposition
when used sarcastically \cite{mohammad2016semeval,augenstein2016stance}.
These cases are difficult for models that rely primarily on surface cues or
single-pass interpretations.

Large language models (LLMs) have demonstrated strong few-shot and
zero-shot performance on stance detection
\cite{brown2020language,wei2022chain,kojima2022zeroshotcot}. However,
single-pass prompting produces only one interpretation of an input, making
predictions brittle when stance is implicit, ambiguous, or rhetorically
complex. Aggregation methods such as self-consistency
\cite{wang2023selfconsistency} can improve robustness by combining multiple
outputs, but they typically aggregate at the label level, discarding the
intermediate reasoning that could help resolve conflicting interpretations.

Recent work on structured reasoning and multi-agent collaboration suggests
a promising alternative. Reasoning-oriented prompting methods, including
Chain-of-Thought \cite{wei2022chain}, Least-to-Most prompting
\cite{zhou2023least2most}, Tree-of-Thoughts \cite{yao2023tree}, and
Graph-of-Thoughts \cite{besta2024graph}, decompose complex problems into
intermediate reasoning steps. Multi-agent frameworks further encourage
diverse and complementary reasoning paths by assigning different agents to
distinct roles or perspectives
\cite{du2024improving,lan2024stance,wu2024autogen,li2023camel,hong2023metagpt}.
However, most existing approaches either aggregate final labels or use fixed
reasoning structures that do not adapt reasoning effort to instance-level
complexity.

We propose \textbf{SMART-D}, a structured multi-agent reasoning framework
for stance detection that performs \emph{reasoning-level synthesis} rather
than label-level voting. SMART-D uses a Manager--Worker architecture in
which Worker agents analyze the input from complementary perspectives and
produce reasoning-only explanations without emitting stance labels. A
Manager agent then synthesizes these explanations to produce the final
stance prediction. To avoid applying the same reasoning budget to every
instance, the Manager dynamically allocates the number of Workers based on
input complexity, assigning more reasoning capacity to ambiguous or
rhetorically complex cases and fewer Workers to more explicit cases.
We evaluate SMART-D on three stance detection benchmarks:
SemEval-2016 \cite{mohammad2016semeval}, P-Stance \cite{li2021p}, and
COVID-19 Stance \cite{glandt2021covidstance}, using three LLMs:
Llama, Mistral, and Gemini. Our results
show that SMART-D provides the largest gains on implicit and
context-dependent stance cases, particularly on COVID-19 Stance, while
remaining competitive on datasets with more explicit stance signals such as
P-Stance.

Our main contributions are as follows:
\begin{itemize}
    \item We introduce SMART-D, a Manager--Worker framework that shifts
    stance aggregation from final-label voting to reasoning-level synthesis.

    \item We propose dynamic Worker allocation, allowing the model to adapt
    reasoning depth to input complexity rather than using a fixed number of
    reasoning paths for all instances.

    \item We provide empirical evidence across three stance benchmarks and
    three LLM backbones showing that adaptive reasoning-level aggregation is
    most beneficial for implicit, ambiguous, and rhetorically complex stance
    cases.
\end{itemize}

\section{Related Work}
\label{sec:related_work}

\paragraph{Stance Detection and LLM Reasoning.}

Stance detection is commonly formulated as a target-conditioned text
classification task
\cite{somasundaran2009stance,mohammad2016semeval,sobhani2016detecting}.
Early work used lexical, syntactic, and task-specific features
\cite{sobhani2016detecting,hasan2013stance}, while recent models improved target-aware representation learning
\cite{augenstein2016stance,zarrella2016mitre,du2017stance,
devlin2019bert,allaway2020zero,liang2022contrastive}. Recent datasets have
expanded stance detection to political and public-health domains
\cite{li2021p,glandt2021covidstance,conforti2020will}, where implicit
stance, domain shift, and target dependence remain persistent challenges
\cite{hardalov2022survey}.

LLMs provide a promising direction for such cases because they can generate
intermediate reasoning before prediction. Chain-of-Thought prompting,
self-consistency, Least-to-Most prompting, Tree-of-Thoughts, and
Graph-of-Thoughts show that decomposed or sampled reasoning can improve
performance on complex tasks
\cite{wei2022chain,kojima2022zeroshotcot,wang2023selfconsistency,
zhou2023least2most,yao2023tree,besta2024graph}. In stance detection,
recent methods incorporate reasoning, verification, rationale generation,
or expert modeling to improve robustness and interpretability
\cite{taranukhin2024stancereasoner,zhang2026msme,wang2024deem,
yan2025collaborative,garg2024stanceformer}. However, many approaches still
aggregate final predictions or use fixed reasoning structures, limiting
their ability to compare competing interpretations explicitly.

\paragraph{Multi-Agent Reasoning and Adaptive Inference.}

Multi-agent LLM frameworks improve robustness by distributing reasoning
across multiple agents or perspectives. Debate-based methods explore
competing interpretations \cite{du2024improving}, collaborative systems
assign specialized roles to agents \cite{lan2024stance}, and general
frameworks such as AutoGen, CAMEL, and MetaGPT support coordinated agent
interaction \cite{wu2024autogen,li2023camel,hong2023metagpt}. Related ideas
also appear in Mixture-of-Experts and dynamic routing models, which route
inputs to specialized modules for capacity and efficiency
\cite{shazeer2017moe,lepikhin2021gshard,du2021glam}. In stance detection,
multi-expert methods such as MSME and DEEM apply expert selection or
routing mechanisms \cite{zhang2026msme,wang2024deem}.

These approaches motivate adaptive computation, but most route at the
representation, expert, or prediction level, rather than at the reasoning
level. Similarly, inference-time methods such as prompt optimization and
self-consistency show that performance depends on
reasoning depth, sampling, and computational budget
\cite{zhou2023prompteng,pryzant2023promptopt,wang2023selfconsistency},
but typically apply a fixed configuration across all inputs.

SMART-D differs by treating reasoning itself as the object of aggregation
and allocation. Worker agents produce reasoning-only explanations from
complementary views, while a Manager synthesizes these explanations
to predict the final stance label. The Manager also dynamically allocates
Workers based on input complexity, assigning more reasoning capacity to
implicit or rhetorically complex cases and fewer Workers to more explicit
ones.

\section{Methodology}
\label{sec:method}

\subsection{Overview}
\label{sec:overview}

We formulate stance detection as a target-conditioned classification task
\cite{mohammad2016semeval}. Given an input text $x$, a stance target $t$, and
a dataset identifier $d$, the goal is to predict a label
$y \in \mathcal{Y}_d$, where $\mathcal{Y}_d$ is the dataset-specific label
space (e.g., {\it favor/against} or {\it favor/against/none}).

SMART-D organizes LLM calls into a structured Manager--Worker architecture.
Unlike label-level aggregation methods, such as majority voting,
self-consistency, or debate-based selection
\cite{wang2023selfconsistency,du2024improving,lan2024stance}, SMART-D
aggregates intermediate explanations. Worker agents
analyze the input from complementary perspectives and produce concise,
evidence-grounded reasoning without emitting stance labels. A Manager agent
then synthesizes these explanations to produce the final prediction. This design is motivated by prior work showing that structured intermediate
reasoning can improve robustness on complex reasoning tasks
\cite{wang2023selfconsistency,yao2023tree,besta2024graph}.

\begin{figure*}[t]
\centering
\includegraphics[width=0.8\textwidth]{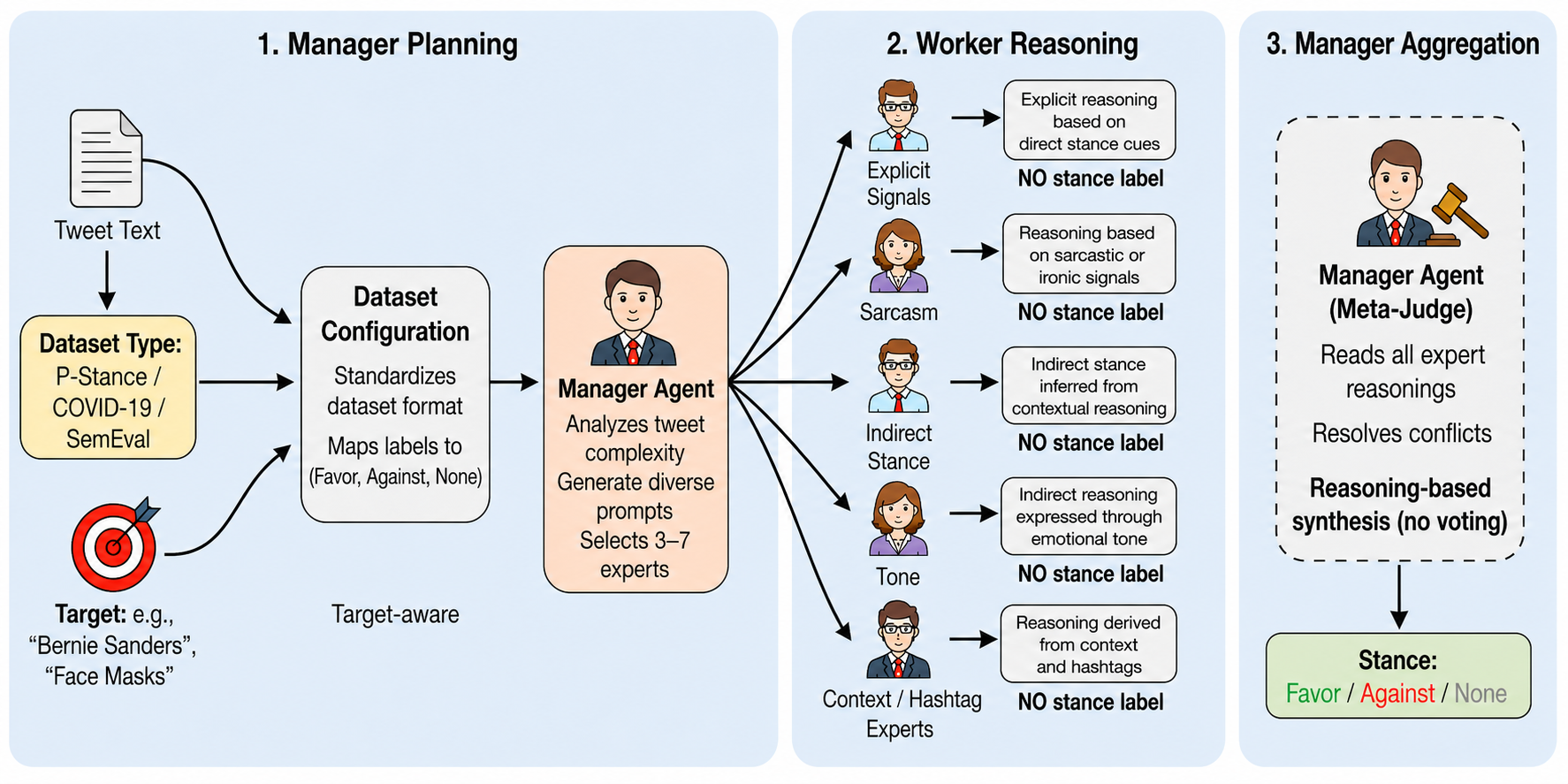}
\caption{
Overview of the SMART-D framework.
The Manager analyzes the input and dynamically allocates Worker agents with
complementary reasoning roles. Workers produce reasoning-only outputs without
emitting stance labels, and the Manager synthesizes these outputs to produce
the final stance label.
}
\label{fig:framework}
\end{figure*}

Figure~\ref{fig:framework} illustrates the overall workflow: the Manager first
plans the reasoning process and allocates Workers, the Workers generate
reasoning-only explanations, and the Manager performs reasoning-level
aggregation to predict final labels.

\subsection{Manager--Worker Reasoning}
\label{sec:architecture}

SMART-D proceeds in three stages.

\paragraph{1. Manager planning.}
Given $(x,t,d)$, the Manager identifies the dataset-specific label space
$\mathcal{Y}_d$, estimates the reasoning complexity of the instance $x$, selects
the number of Workers, and assigns each Worker a distinct analytical role.
Roles are designed to capture complementary stance signals, such as explicit
stance cues, sarcasm or irony, indirect stance through related entities,
emotional tone, and contextual framing.

\paragraph{2. Worker reasoning.}
Each Worker analyzes the same input $(x,t,d)$ under its assigned perspective.
Workers are constrained to produce short, evidence-grounded explanations and
are explicitly prohibited from outputting stance labels. This
\emph{reasoning-only constraint} prevents premature commitment to a label and
preserves intermediate evidence for downstream synthesis.

\paragraph{3. Manager aggregation.}
The Manager receives all Worker explanations and predicts a single stance
label by comparing the consistency,
relevance, and evidential support of the explanations. This allows the model
to reconcile competing interpretations, which is particularly important when
stance is implicit, ambiguous, or rhetorically framed.

\subsection{Dynamic Worker Allocation}
\label{sec:dynamic_workers}

A fixed number of reasoning paths may be ineffective: too few Workers can miss
important interpretations, while too many can introduce redundancy,
conflicting evidence, and additional cost. This trade-off is consistent with
prior work on multi-path reasoning, where additional reasoning samples do not
always yield monotonic gains \cite{wang2023selfconsistency,yao2023tree}. SMART-D therefore treats reasoning
depth as an adaptive computational resource. For each instance, the Manager
selects $N \in \{3,\ldots,7\}$ Workers based on input complexity. Inputs with
explicit stance cues are assigned fewer Workers, while cases involving
sarcasm, indirect references, multiple entities, or complex rhetorical
structure receive additional  perspectives. 

\subsection{Implementation Details}
\label{sec:implementation}

We implement SMART-D through structured prompts that specify Manager and Worker roles, enforce the reasoning-only constraint, and encourage complementary reasoning perspectives. The Manager operates in two prompt stages: a planning stage that allocates Workers and a decision stage that synthesizes Worker explanations into a final label. All experiments are conducted in a zero-shot prompting setting without task-specific fine-tuning or in-context examples. All LLM inference was performed through the OpenRouter API using a unified OpenAI-compatible interface. Workers use a higher temperature ($T = 0.7$) to encourage diverse reasoning, while the Manager uses a lower temperature ($T = 0.3$) to promote stable aggregation. All other decoding parameters are kept at their default values unless otherwise specified. Full prompt templates used for Manager/Workers are provided in Appendix~\ref{app:prompts}.

The workflow logs selected Worker counts, Worker roles, generated explanations,  and final predictions for reproducibility and qualitative analysis. 
 All prompts, configurations, and evaluation
scripts will be released to support reproducibility.

\section{Experimental Setup}
\label{sec:exp}

\subsection{Models}
\label{sec:models}

We evaluate SMART-D using three LLM backbones:
\textbf{Gemini-3-Flash-preview}, \textbf{Llama-3.3 70B}, and
\textbf{Mistral-3.2 24B}. For consistency, we refer to these models as Gemini, Llama, and Mistral throughout the paper. These models span different parameter
scales and capability profiles, allowing us to assess whether
reasoning-level aggregation generalizes across heterogeneous LLMs.
For each experiment, the same LLM is used for both the Manager
and Worker agents. Manager and Worker behaviors differ only through
role-specific prompts and decoding settings, rather than through different
model architectures or fine-tuning. This design isolates the effect of the
SMART-D reasoning framework from heterogeneous-model effects, which we leave
to future work.

\subsection{Datasets}
\label{sec:datasets}

We evaluate on three stance detection benchmarks:
SemEval-2016 Task 6 \cite{mohammad2016semeval}, P-Stance \cite{li2021p},
and COVID-19 Stance \cite{glandt2021covidstance}. These datasets cover both
political and public-health domains and include varying degrees of implicit
stance, making them suitable for evaluating reasoning-oriented approaches. Some representative dataset examples and target-label pairs are provided in
Appendix~\ref{app:dataset_examples}.

\paragraph{SemEval-2016 Task 6.}
A target-conditioned dataset with five targets: Atheism, Climate
Change, Feminist Movement, Hillary Clinton, and Legalization of Abortion.
Labels are {\it favor}, {\it against}, {\it none}.

\paragraph{P-Stance.}
A political stance dataset with three U.S.\ presidential targets: Donald
Trump, Joe Biden, and Bernie Sanders. The task is binary text classification with
{\it favor} and {\it against} labels.

\paragraph{COVID-19 Stance.}
A public-health stance dataset covering four targets:
Anthony Fauci, face masks, stay-at-home orders, and school closures. Labels
are {\it favor}, {\it against}, {\it none}.

\subsection{Evaluation Protocol}
\label{sec:metrics}

We report \textbf{Macro-F1} and \textbf{Accuracy}. Macro-F1 is the primary
metric because stance datasets often exhibit class imbalance and require
robust performance across minority classes. 
Each experiment is repeated three times, and we report the mean and standard
deviation. 

\begin{table*}[!t]
\centering
\begingroup
\fontsize{7.8}{10.2}\selectfont
\begin{adjustbox}{max width=\textwidth}
\begin{tabular}{llccccc}
\toprule
\multirow{2}{*}{Category} & \multirow{2}{*}{Model} & \multicolumn{5}{c}{COVID-19 (\%)} \\
\cmidrule(lr){3-7}
 &  & FM & AF & SH & SC & Avg $\pm$ std \\
\midrule

\multirow{3}{*}{Supervised baselines (prior work)}
& BERT     & 80.3 & 81.8 & 80.0 & 75.5 & 79.4 \\
& BERT-NS  & 83.3 & 82.1 & 78.4 & 75.3 & 79.8 \\
& BERT-DAN & 82.5 & \textbf{83.2} & 78.7 & 71.7 & 79.0 \\

\midrule

\multirow{3}{*}{Baseline (Single-Model)}
& Gemini
 & \underline{88.11} & 79.98 & \underline{84.21} & 76.01 & \underline{82.08 $\pm$ 0.05} \\
& Llama
 & 82.67 & 77.14 & 81.08 & 38.23 & 69.78 $\pm$ 0.29 \\
& Mistral
 & 79.85 & 74.95 & 75.16 & 41.20 & 67.79 $\pm$ 0.17 \\

\midrule

\multirow{3}{*}{SMART-D (ours)}
& Gemini
 & \textbf{88.37} & 82.52 & \textbf{86.03} & \textbf{87.35} & \textbf{86.07 $\pm$ 0.38} \\
& Llama
 & 84.60 & 79.49 & 78.68 & \underline{82.47} & 81.31 $\pm$ 0.88  \\
& Mistral
 & 81.89 & 76.56 & 75.03 & 70.49 & 75.99 $\pm$ 0.33 \\

\bottomrule
\end{tabular}
\end{adjustbox}
\endgroup
\vspace{-2mm}
\caption{Macro-F1 results on the COVID-19 Stance dataset across four targets:
Wearing a Face Mask (FM), Anthony Fauci (AF), Stay-at-Home Orders (SH), and
Keeping Schools Closed (SC). Supervised baselines are reported from
\citet{glandt2021covidstance}. Baseline (Single-Model) and SMART-D results
are computed in our experimental setup; for these methods, results are
reported as mean $\pm$ standard deviation over three runs. Best and
second-best results in each column are shown in \textbf{bold} and underlined,
respectively.}
\label{tab:covid_f1}
\end{table*}
\section{Results and Discussion}
\label{sec:results}

We evaluate SMART-D  using Macro-F1, with accuracy reported in
Appendix~\ref{sec:accuracy_results}. Results for prior methods are taken from
published work, while single-model baselines are computed using our own
implementation with the same LLM backbones.

\subsection{Main Results}
\label{sec:results_f1}

Table~\ref{tab:covid_f1} shows  Macro-F1 results for COVID-19 Stance and Table~\ref{tab:f1} shows results for SemEval-2016 and P-Stance. Overall, SMART-D provides the
largest gains on COVID-19 Stance, a dataset with many implicit and
context-dependent stance expressions, while achieving smaller but positive
 gains on SemEval-2016 and mixed results on P-Stance.

\paragraph{COVID-19 Stance.}
On COVID-19 Stance, SMART-D improves average Macro-F1 for all three
backbones. With Gemini, SMART-D achieves the best overall
score of \textbf{86.07}, improving over the corresponding single-model
baseline by \textbf{+3.99} points and outperforming the strongest supervised
baseline by more than \textbf{+6} points. The gains are especially large 
for \emph{School Closures}, where Gemini improves from 76.01 to
\textbf{87.35}. Llama also shows a substantial average improvement,
from 69.78 to \textbf{81.31}. While Llama exhibits a sharp performance drop on the Stay-at-Home (SH) target (38.23 Macro-F1)—likely struggling to map indirect policy references to the correct stance—SMART-D's structured reasoning paths successfully resolve this ambiguity. These results support our main hypothesis: reasoning-level aggregation is most useful when stance depends on rhetorical framing, indirect references,
or target-specific interpretation. The improvements over the corresponding
single-model baselines are stable across repeated runs; detailed statistical
significance tests are reported in Appendix~\ref{app:statistical_significance}.


\begin{table*}[!t]
\centering
\begingroup
\fontsize{8.6}{10.2}\selectfont
\resizebox{\textwidth}{!}{%
\begin{tabular}{llcccccc|cccc}
\toprule
\multirow{2}{*}{Category} & \multirow{2}{*}{Model} &
\multicolumn{6}{c|}{SemEval16 (\%)} &
\multicolumn{4}{c}{P-Stance (\%)} \\
\cmidrule(lr){3-8}\cmidrule(lr){9-12}
 &  & A & CC & FM & HC & LA & Avg $\pm$ std
 & DT & JB & BS & Avg $\pm$ std \\
\midrule
\multirow{6}{*}{Zero-shot (LLMs)} &
Base
(GPT-3.5)
 & 58.30 & 51.10 & 62.30 & 65.00 & 60.80 & 59.50
 & 67.30 & 78.20 & 71.60 & 72.40 \\
&
CoT
 & 64.10 & 55.70 & 62.40 & 70.70 & 61.90 & 63.20
 & 71.40 & 80.50 & 74.10 & 75.30 \\
&
BKEL
 & 71.50 & 66.00 & 63.10 & 76.50 & 64.20 & 68.30
 & 80.30 & 78.30 & 79.60 & 79.40 \\
&
Stance Reasoner
 & 69.70 & 62.50 & 73.90 & 67.70 & 60.30 & 66.80
 & 79.50 & 81.00 & 79.60 & 80.00 \\
&
COLA
 & 70.80 & 65.50 & 63.40 & 81.70 & 71.00 & 70.50
 & 86.60 & 84.00 & 79.70 & 83.40 \\
&
ToC
 & -- & -- & -- & -- & -- & 69.40
 & 75.70 & 83.10 & 80.40 & 79.70 \\
\midrule
\multirow{4}{*}{MSME} &
GPT-3.5
 & 75.20 & 74.90 & 72.50 & 81.10 & 69.90 & 74.70
 & 87.70 & 84.90 & 82.80 & 85.10 \\
&
GPT-4o
 & 80.30 & 76.20 & 75.50 & 81.90 & 71.90 & 77.20
 & 88.60 & 85.60 & \underline{84.10} & \underline{86.10} \\
&
DeepSeek-r1
 & 81.50 & 78.50 & 74.80 & 80.60 & 73.50 & 77.80
 & 87.10 & 84.70 & \textbf{84.50} & 85.40 \\
&
QwQ-32B
 & 79.50 & 77.10 & 76.30 & 76.90 & 68.40 & 75.60
 & 85.10 & 84.30 & 83.50 & 84.30 \\
\midrule
\multirow{3}{*}{Baseline (Single-Model)} &
Gemini
 & 76.71 & 85.58 & 79.55 & 89.18 & 69.46 & 80.10 $\pm$ 0.06
 & \textbf{92.56} & \underline{86.28} & 84.07 & \textbf{87.64 $\pm$ 0.00} \\
&
Llama
 & 77.30 & 87.22 & 77.03 & 89.19 & 70.43 & 80.23 $\pm$ 0.07
 & 89.35 & \textbf{86.85} & 83.06 & 86.42 $\pm$ 0.13 \\
&
Mistral
 & 69.17 & 71.77 & \underline{80.40} & 88.09 & \underline{76.46} & 77.18 $\pm$ 0.05
 & 83.20 & 83.27 & 79.11 & 81.86 $\pm$ 0.46 \\
\midrule
\multirow{3}{*}{SMART-D (Ours)} &
Gemini
 & \textbf{87.75} & \underline{88.16} & \textbf{81.00} & 86.50 & 71.13 & \underline{82.77 $\pm$ 0.74}
 & \underline{92.39} & 85.49 & 82.66 & \underline{86.84 $\pm$ 0.37} \\
&
Llama
 & \underline{84.74} & \textbf{91.52} & 75.40 & \textbf{89.56} & 73.29 & \textbf{82.90 $\pm$ 0.26}
 & 89.47 & 85.73 & 82.69 & 85.96 $\pm$ 0.24 \\
&
Mistral
 & 75.21 & 83.13 & 78.44 & \underline{89.33} & \textbf{77.67} & 80.76 $\pm$ 0.45
 & 89.22 & 85.23 & 81.99 & 85.48 $\pm$ 0.15 \\
\bottomrule
\end{tabular}%
}
\endgroup
\vspace{-2mm}
\caption{
Macro-F1 results on SemEval-2016 Task 6 \cite{mohammad2016semeval}
and P-Stance \cite{li2021p}. Prior zero-shot and MSME results are reported
from \citet{zhang2026msme}; Baseline (Single-Model) and SMART-D results are
computed in our experimental setup. SemEval-2016 targets: A = Atheism,
CC = Climate Change, FM = Feminist Movement, HC = Hillary Clinton,
LA = Legalization of Abortion. P-Stance targets: DT = Donald Trump,
JB = Joe Biden, BS = Bernie Sanders. For our methods, results are reported
as mean $\pm$ standard deviation over three runs. Best and second-best
results in each column are shown in \textbf{bold} and underlined,
respectively.
}
\label{tab:f1}
\end{table*}

\paragraph{SemEval-2016.}
On SemEval-2016, SMART-D improves average Macro-F1 for all three backbones.
The best result is obtained with Llama, reaching
\textbf{82.90}, a \textbf{+2.67} point improvement over its single-model
baseline. Gemini and Mistral also improve in average Macro-F1, although the
target-level effects are not uniform. SMART-D also surpasses strong prior methods such as MSME, showing that reasoning-level aggregation provides benefits beyond standard prompting or multi-expert strategies.
These results suggest that  SMART-D is
beneficial on SemEval-2016, but the gains are more moderate than on
COVID-19, consistent with a dataset containing both implicit and more
lexically explicit stance expressions.

\paragraph{P-Stance.}
On P-Stance, SMART-D remains competitive with strong prior LLM-based methods,
but its gains over single-model baselines are model-dependent. Mistral
benefits substantially, improving from 81.86 to \textbf{85.48} Macro-F1. In
contrast, Gemini and Llama decrease slightly relative
to their single-model baselines. This indicates that additional reasoning is
not uniformly beneficial when stance is already expressed explicitly. Rather,
SMART-D is most effective when the input requires ambiguity resolution or
pragmatic interpretation.

Taken together, these results show that SMART-D is strongest in the setting
for which it was designed: implicit, ambiguous, and rhetorically complex
stance detection. On datasets with more explicit stance cues, the framework
remains competitive, but the benefit of additional reasoning depends on the
base model and input complexity.

\subsection{Ablation and Allocation Analysis}
\label{sec:worker_analysis}

We analyze SMART-D through ablations and diagnostic allocation studies. We
first examine how Worker count affects performance, then evaluate the role of
reasoning-level synthesis, dynamic allocation behavior, and cross-model Worker
selection.

\paragraph{Effect of the number of Workers.}
We first evaluate fixed Worker configurations with
$N \in \{1,3,5,7,9,11\}$ and compare them with dynamic allocation.
Table~\ref{tab:ablation_workers_all_full} in Appendix~\ref{app:worker_count_full} reports Macro-F1 across all datasets and backbones. For brevity, Table~\ref{tab:worker_summary}
summarizes the best fixed configuration among
$N \in \{1,3,5,7,9,11\}$ and compares it with dynamic allocation. Overall, performance does not increase monotonically with more Workers. Larger $N$ values  can improve performance, but the gains are inconsistent and come with higher inference cost. This suggests a trade-off between \emph{reasoning diversity} and \emph{aggregation noise}: additional Workers provide more perspectives, but can also introduce redundant or conflicting explanations.

The dynamic strategy is constrained to $N \in \{3,\cdots,7\}$ to reflect a
practical reasoning budget. While it is not always the best-performing
configuration, it provides a strong compromise between performance and cost.
On COVID-19, dynamic allocation performs especially well, achieving the best score for Gemini and near-best performance for Llama.
On P-Stance and SemEval-2016, fixed larger budgets sometimes perform better,
suggesting that the optimal amount of reasoning depends on both dataset
and backbone. Detailed analyses of token cost are provided in Appendix~\ref{app:tradeoff}.

\begin{figure*}[!t]
\centering
\includegraphics[width=0.8\linewidth]{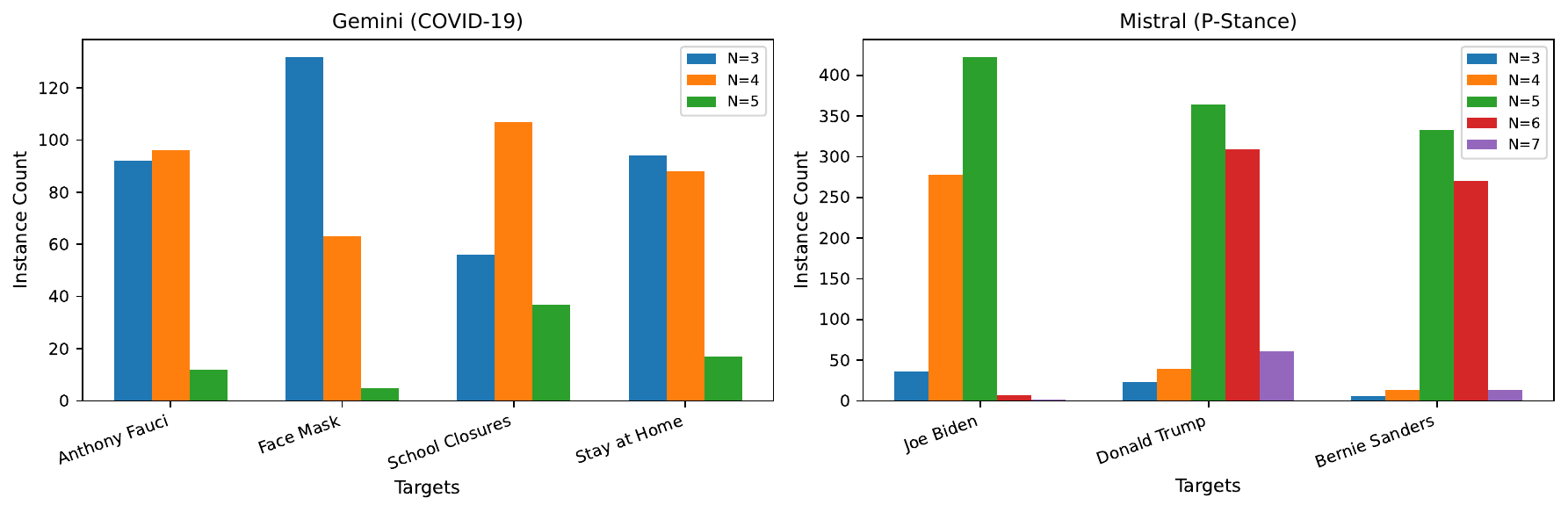}
\caption{
Distribution of dynamically selected Worker counts for two representative
settings. Left: COVID-19 with Gemini, which yields the lowest
average allocation ($\bar{N}=3.60$). Right: P-Stance with Mistral,
which yields the highest average allocation ($\bar{N}=5.13$). Allocation
patterns vary by both dataset and backbone model.
}
\label{fig:worker_distribution}
\end{figure*}

\paragraph{Dynamic allocation behavior.}
To examine how the Manager uses the dynamic allocation mechanism, we analyze
the distribution of selected Worker counts for two representative settings:
Gemini on COVID-19, which has the lowest average allocation
($\bar{N}=3.60$), and Mistral on P-Stance, which has the highest
average allocation ($\bar{N}=5.13$). Figure~\ref{fig:worker_distribution}
shows that allocation is structured rather than uniform. In the COVID-19
setting, the Manager primarily selects smaller Worker counts ($N=3$--$4$),
whereas in the P-Stance setting selections shift toward larger values
($N=5$--$6$).

This pattern suggests that Worker allocation depends not only on dataset-level
difficulty, but also on the backbone model and its uncertainty under the
Manager prompt. In particular, the strong Gemini backbone can often resolve
COVID-19 cases with fewer Workers, whereas the smaller Mistral backbone tends
to request more reasoning perspectives on P-Stance. Thus, dynamic allocation
should be interpreted as an instance- and model-dependent estimate of
reasoning need, rather than a direct proxy for dataset difficulty. Detailed per-target allocation statistics and planning-stage examples are provided in Appendix~\ref{app:additional}.

\begin{table}[t]
\centering
\fontsize{7.2}{8.8}\selectfont
\setlength{\tabcolsep}{3.5pt}
\begin{tabular}{llcccc}
\toprule
\textbf{Dataset} & \textbf{Model} & \textbf{Best fixed} & \textbf{$N$} & \textbf{Dyn.} & $\bar{N}$ \\
\midrule
\multirow{3}{*}{COVID-19}
& Gemini  & 85.64 & 7  & \textbf{86.07} & 3.60 \\
& Llama   & \textbf{81.46} & 11 & 81.31 & 4.83 \\
& Mistral & \textbf{77.09} & 7  & 75.99 & 4.82 \\
\midrule
\multirow{3}{*}{P-Stance}
& Gemini  & \textbf{87.19} & 11 & 86.84 & 4.54 \\
& Llama   & \textbf{86.27} & 7  & 85.96 & 5.00 \\
& Mistral & \textbf{85.71} & 11 & 85.48 & 5.13 \\
\midrule
\multirow{3}{*}{SemEval16}
& Gemini  & \textbf{82.97} & 5  & 82.77 & 4.68 \\
& Llama   & \textbf{84.92} & 11 & 82.90 & 4.82 \\
& Mistral & \textbf{81.96} & 7  & 80.76 & 4.00 \\
\bottomrule
\end{tabular}
\caption{
Worker-count ablation summary (Macro-F1). ``Best fixed'' denotes the
best fixed-$N$ configuration among $N \in \{1,3,5,7,9,11\}$. Dynamic
allocation uses $N \in \{3,\ldots,7\}$. For the dynamic
 setting, $\bar{N}$ denotes the average number of selected Workers. The
 best result in each row is shown in \textbf{bold}.
}
\label{tab:worker_summary}
\end{table}

\paragraph{Effect of reasoning-level synthesis.}
We next ablate the reasoning components on COVID-19, the benchmark where
SMART-D shows the largest gains. Table~\ref{tab:ablation_full} compares
Worker voting (WV), Worker reasoning (WR), Manager reasoning (MR),
self-consistency, and the full SMART-D framework. In WV settings, Workers
directly output stance labels and aggregation is performed over labels. In
WR settings, Workers instead produce intermediate explanations. MR denotes
explicit Manager-side synthesis before the final prediction.

The results show that Worker-level reasoning is the largest contributor to
performance. For Gemini, adding Worker reasoning improves
Macro-F1 from 78.95 to 85.27, substantially outperforming both Worker voting
and self-consistency. Similar trends hold for Llama and
Mistral. Adding Manager reasoning provides smaller additional gains,
indicating that the main benefit comes from exposing the Manager to richer
intermediate explanations rather than only final labels. The full SMART-D framework achieves the best result for each backbone,
confirming that dynamic allocation and reasoning-level aggregation are
complementary. Full per-target ablation results are provided in
Appendix~\ref{app:per_target_ablation}. 

\begin{table}[!t]
\centering
\fontsize{7.8}{9.5}\selectfont
\setlength{\tabcolsep}{4pt}
\begin{tabular}{l l c}
\toprule
\textbf{Method} & \textbf{Model} & \textbf{Avg $\pm$ std} \\
\midrule
WV + MR & Gemini & 78.45 $\pm$ 0.52 \\
 & Llama & 57.59 $\pm$ 1.18 \\
 & Mistral & 60.02 $\pm$ 1.71 \\
\midrule
WV & Gemini & 78.95 $\pm$ 1.41 \\
 & Llama & 66.18 $\pm$ 0.78 \\
 & Mistral & 62.20 $\pm$ 1.27 \\
\midrule
WV + WR & Gemini & 85.27 $\pm$ 0.62 \\
 & Llama & 72.99 $\pm$ 0.72 \\
 & Mistral & 71.17 $\pm$ 0.52 \\
\midrule
WV + WR + MR & Gemini & 85.00 $\pm$ 0.18 \\
 & Llama & 75.10 $\pm$ 0.28 \\
 & Mistral & 72.35 $\pm$ 0.31 \\
\midrule
Self-Consistency & Gemini & 82.36 $\pm$ 0.18 \\
 & Llama & 70.61 $\pm$ 0.28 \\
 & Mistral & 68.67 $\pm$ 0.33 \\
\midrule
\textbf{SMART-D} & Gemini & \textbf{\underline{86.07 $\pm$ 0.38}} \\
 & Llama & \textbf{81.31 $\pm$ 0.88} \\
 & Mistral & \textbf{75.99 $\pm$ 0.33} \\
\bottomrule
\end{tabular}
\caption{
Ablation of reasoning components on the COVID-19 dataset (Macro-F1, \%).
WV: Worker Voting; WR: Worker Reasoning; MR: Manager Reasoning.
Self-consistency aggregates 5 independent generations per run.
Best result for each backbone is shown in \textbf{bold}, and the best result overall is also underlined.
}
\label{tab:ablation_full}
\end{table}

\paragraph{Cross-model Worker allocation.}
We further examine whether heterogeneous Worker pools improve
reasoning-level aggregation. Using COVID-19 Stance as the benchmark where
SMART-D shows the largest gains, we compare the standard homogeneous setting,
where the Manager and Workers use the same backbone, with a mixed-model
setting in which the Manager selects Workers from the set
{Gemini, Llama, and Mistral}.

As shown in Table~\ref{tab:cross_model_allocation}, heterogeneous Workers do
not consistently improve performance. For Gemini and
Llama, the homogeneous SMART-D setting performs better, suggesting
that consistency in reasoning style may be important for stable aggregation.
In contrast, Mistral benefits substantially from mixed-model Workers,
improving from 75.99 to 85.23 Macro-F1. One possible explanation is that a
weaker Manager can benefit from reasoning traces produced by stronger Worker
models, whereas stronger Managers may be more sensitive to inconsistencies
introduced by heterogeneous reasoning styles.
Overall, these results suggest that model diversity alone does not guarantee
better reasoning-level aggregation. Cross-model Worker allocation may be
useful for weaker backbones, but homogeneous reasoning appears more stable
for stronger backbones. Additional Worker selection patterns are provided in
Appendix~\ref{app:worker_model_distribution}.

\begin{table}[t]
\centering
\fontsize{7.2}{8.8}\selectfont
\setlength{\tabcolsep}{4pt}
\begin{tabular}{lcc}
\toprule
\textbf{Manager} & \textbf{Homogeneous} & \textbf{Mixed Workers} \\
\midrule
Gemini & \textbf{86.07 $\pm$ 0.38} & 82.09 $\pm$ 0.67 \\
Llama          & \textbf{81.31 $\pm$ 0.88} & 78.78 $\pm$ 0.81 \\
Mistral         & 75.99 $\pm$ 0.33 & \textbf{85.23 $\pm$ 0.85} \\
\bottomrule
\end{tabular}
\caption{
Cross-model Worker allocation on COVID-19 Stance (Macro-F1, \%).
Homogeneous SMART-D uses the same backbone for Manager and Workers; mixed
Workers are selected from the three-backbone pool.
}
\label{tab:cross_model_allocation}
\end{table}

\subsection{Qualitative Case Studies}
\label{sec:case_study}

To illustrate where SMART-D is most useful, we examine two examples from
COVID-19 Stance for the target \emph{School Closures}. In both cases,
the single-model baseline fails, while SMART-D succeeds. Full reasoning traces are provided in
Appendix~\ref{app:case_studies}.

\begin{casebox}{Case 1: Rhetorical challenge and implicit support}
\small
\textbf{Tweet:}
\emph{``Governor @GovRonDeSantis, why aren't your offices open? If you want
all schools open in a month, open your offices today. \#COVID19''}

\vspace{2pt}
\textbf{Gold:} \labeltag{correctbg}{\it favor}
\textbf{Baseline:} \labeltag{errorbg}{\it against}
\textbf{SMART-D:} \labeltag{correctbg}{\it favor}
\quad
\textbf{Workers:} 5

\vspace{3pt}
\textbf{Why SMART-D succeeds:}
Workers identify the tweet as a rhetorical challenge: reopening schools is
framed as unsafe because government offices remain closed. This requires
reasoning beyond surface-level criticism.
\end{casebox}

\begin{casebox}{Case 2: Distinguishing  stance from related complaints}
\small
\textbf{Tweet:}
\emph{``What if there was job loss or pay cut due to recession? Would
government still give decision to not pay fees to schools? Are schools run by
robots?? Teachers also need to get paid right?''}

\vspace{2pt}
\textbf{Gold:} \labeltag{correctbg}{\it none}
\textbf{Baseline:} \labeltag{errorbg}{\it against}
\textbf{SMART-D:} \labeltag{correctbg}{\it none}
\quad
\textbf{Workers:} 3

\vspace{3pt}
\textbf{Why SMART-D succeeds:}
Workers separate general financial and institutional concerns from a direct
stance on school closures, enabling a target-aware {\it none} prediction.
\end{casebox}

These cases show that SMART-D is especially useful when stance depends on
rhetorical framing,  or on distinguishing the target from
related issues.

\subsection{Discussion}
\label{sec:discussion}

Overall, the results indicate that SMART-D is most useful when stance
requires contextual interpretation. The largest gains occur on COVID-19
Stance, where many examples involve sarcasm, rhetorical comparison, policy
framing, or indirect references. Gains are more moderate on SemEval-2016 and
more model-dependent on P-Stance, suggesting that additional reasoning is most
valuable when stance is implicit or ambiguous, but less useful when stance is
expressed through direct lexical cues.
The ablations support this interpretation. Increasing the number of Workers
does not monotonically improve performance, reflecting a trade-off between
reasoning diversity and aggregation noise. Dynamic allocation helps manage
this trade-off by assigning additional reasoning capacity selectively rather
than applying a fixed budget to all instances. Similarly, the cross-model
results show that diversity alone is not sufficient: effective aggregation
depends on both the strength and compatibility of Worker reasoning traces.

Qualitative examples reinforce the quantitative findings: SMART-D succeeds
when Workers capture complementary interpretations and the Manager resolves
them under the target-specific label space. Thus, SMART-D should be viewed as
an adaptive reasoning framework for complex stance cases, rather than a
uniformly better inference strategy for all instances.

\section{Conclusion}
\label{sec:conclusion}

We introduced SMART-D, a structured multi-agent reasoning framework for
target-aware stance detection. SMART-D decomposes stance analysis into
complementary reasoning-only Worker perspectives and uses a Manager to
synthesize these explanations into a final stance prediction. The framework
also adapts inference-time reasoning effort by dynamically selecting the
number of Workers based on input complexity.

Experiments on SemEval-2016, P-Stance, and COVID-19 Stance show that
SMART-D achieves its largest gains on implicit, context-dependent, and
rhetorically complex stance cases, while gains are smaller and more
model-dependent when stance cues are explicit. These findings suggest that
reasoning-level aggregation is most useful when stance cannot be reliably
inferred from surface cues alone, and highlight adaptive reasoning as a
promising direction for robust real-world stance detection.

\clearpage
\section{Limitations}
\label{sec:limitations}

This study has several limitations. First, our experiments are limited to
English-language Twitter/X datasets, and the effectiveness of SMART-D on
other languages, platforms, or longer-form texts remains an open question.
Second, the multi-agent reasoning architecture introduces additional
computational overhead compared to single-pass prompting. Although dynamic
Worker allocation reduces unnecessary computation, SMART-D still requires
multiple reasoning passes per instance.
Third, the benefits of structured reasoning are condition-dependent. Gains
are most pronounced on datasets with implicit or complex stance expressions,
such as COVID-19 Stance, while improvements are more limited on datasets with
more explicit stance signals, such as P-Stance. Finally, models applied to social media data may reflect biases in
those datasets, including demographic, ideological, or platform-specific
imbalances.

From an ethical perspective, stance detection systems can support large-scale
analysis of public discourse, but they may also be misused to monitor,
influence, or manipulate online debates. Responsible deployment should
emphasize transparency, careful interpretation, and safeguards against uses
that distort public opinion.

\bibliography{custom,anthology}

@inproceedings{mohammad2016semeval,
  title={Semeval-2016 task 6: Detecting stance in tweets},
  author={Mohammad, Saif and Kiritchenko, Svetlana and Sobhani, Parinaz and Zhu, Xiaodan and Cherry, Colin},
  booktitle={Proceedings of the 10th international workshop on semantic evaluation (SemEval-2016)},
  pages={31--41},
  year={2016}
}

@inproceedings{hasan2013stance,
  title={Stance classification of ideological debates: Data, models, features, and constraints},
  author={Hasan, Kazi Saidul and Ng, Vincent},
  booktitle={Proceedings of the sixth international joint conference on natural language processing},
  pages={1348--1356},
  year={2013}
}

@inproceedings{augenstein2016stance,
  title={Stance detection with bidirectional conditional encoding},
  author={Augenstein, Isabelle and Rockt{\"a}schel, Tim and Vlachos, Andreas and Bontcheva, Kalina},
  booktitle={Proceedings of the 2016 conference on empirical methods in natural language processing},
  pages={876--885},
  year={2016}
}

@inproceedings{zarrella2016mitre,
  title={Mitre at semeval-2016 task 6: Transfer learning for stance detection},
  author={Zarrella, Guido and Marsh, Amy},
  booktitle={Proceedings of the 10th international workshop on semantic evaluation (SemEval-2016)},
  pages={458--463},
  year={2016}
}

@inproceedings{sobhani2016detecting,
  title={Detecting stance in tweets and analyzing its interaction with sentiment},
  author={Sobhani, Parinaz and Mohammad, Saif and Kiritchenko, Svetlana},
  booktitle={Proceedings of the fifth joint conference on lexical and computational semantics},
  pages={159--169},
  year={2016}
}

@inproceedings{du2017stance,
  title={Stance classification with target-specific neural attention networks},
  author={Du, Jiachen and Xu, Ruifeng and He, Yulan and Gui, Lin},
  booktitle={26th International Joint Conference on Artificial Intelligence, IJCAI 2017},
  pages={3988--3994},
  year={2017},
  organization={International Joint Conferences on Artificial Intelligence}
}

@inproceedings{devlin2019bert,
  title={Bert: Pre-training of deep bidirectional transformers for language understanding},
  author={Devlin, Jacob and Chang, Ming-Wei and Lee, Kenton and Toutanova, Kristina},
  booktitle={Proceedings of the 2019 conference of the North American chapter of the association for computational linguistics: human language technologies, volume 1 (long and short papers)},
  pages={4171--4186},
  year={2019}
}

@inproceedings{allaway2020zero,
  title={Zero-shot stance detection: A dataset and model using generalized topic representations},
  author={Allaway, Emily and McKeown, Kathleen},
  booktitle={Proceedings of the 2020 Conference on Empirical Methods in Natural Language Processing (EMNLP)},
  pages={8913--8931},
  year={2020}
}

@inproceedings{conforti2020will,
  title={Will-they-won’t-they: A very large dataset for stance detection on Twitter},
  author={Conforti, Costanza and Berndt, Jakob and Pilehvar, Mohammad Taher and Giannitsarou, Chryssi and Toxvaerd, Flavio and Collier, Nigel},
  booktitle={Proceedings of the 58th annual meeting of the association for computational linguistics},
  pages={1715--1724},
  year={2020}
}

@article{brown2020language,
  title={Language models are few-shot learners},
  author={Brown, Tom and Mann, Benjamin and Ryder, Nick and Subbiah, Melanie and Kaplan, Jared D and Dhariwal, Prafulla and Neelakantan, Arvind and Shyam, Pranav and Sastry, Girish and Askell, Amanda and others},
  journal={Advances in neural information processing systems},
  volume={33},
  pages={1877--1901},
  year={2020}
}

@inproceedings{glandt2021covidstance,
  title={Stance Detection in COVID-19 Tweets},
  author={Glandt, Kyle and Khanal, Sarthak and Li, Yingjie and Caragea, Doina and Caragea, Cornelia},
  booktitle={Proceedings of the 59th Annual Meeting of the Association for Computational Linguistics (ACL)},
  year={2021},
  pages={1596--1611}
}

@inproceedings{li2021p,
  title={P-stance: A large dataset for stance detection in political domain},
  author={Li, Yingjie and Sosea, Tiberiu and Sawant, Aditya and Nair, Ajith Jayaraman and Inkpen, Diana and Caragea, Cornelia},
  booktitle={Findings of the association for computational linguistics: ACL-IJCNLP 2021},
  pages={2355--2365},
  year={2021}
}

@inproceedings{hardalov2022survey,
  title={A Survey on Stance Detection for Mis- and Disinformation Identification},
  author={Hardalov, Momchil and Arora, Arnav and Nakov, Preslav and Augenstein, Isabelle},
  booktitle={Findings of the Association for Computational Linguistics: NAACL 2022},
  year={2022}
}

@inproceedings{liang2022contrastive,
  author = {Liang, Bin and Chen, Zixiao and Gui, Lin and He, Yulan and Yang, Min and Xu, Ruifeng},
  title = {Zero-Shot Stance Detection via Contrastive Learning},
  booktitle = {Proceedings of WWW},
  pages = {2738--2747},
  year = {2022}
}

@inproceedings{wang2023selfconsistency,
  title={Self-Consistency Improves Chain of Thought Reasoning in Language Models},
  author={Wang, Xuezhi and Wei, Jason and Schuurmans, Dale and Le, Quoc V. and Chi, Ed H. and Narang, Sharan and Chowdhery, Aakanksha and Zhou, Denny},
  booktitle={International Conference on Learning Representations (ICLR)},
  year={2023}
}

@inproceedings{kojima2022zeroshotcot,
  title={Large Language Models are Zero-Shot Reasoners},
  author={Kojima, Takeshi and Gu, Shixiang Shane and Reid, Machel and Matsuo, Yutaka and Iwasawa, Yusuke},
  booktitle={Advances in Neural Information Processing Systems (NeurIPS)},
  year={2022}
}

@inproceedings{zhou2023least2most,
  title={Least-to-Most Prompting Enables Complex Reasoning in Large Language Models},
  author={Zhou, Denny and Sch{\"a}rli, Nathanael and Hou, Le and Wei, Jason and Scales, Nathan and Wang, Xuezhi and Schuurmans, Dale and Cui, Claire and Bousquet, Olivier and Le, Quoc V and Chi, Ed H},
  booktitle={International Conference on Learning Representations (ICLR)},
  year={2023}
}

@article{yao2023tree,
  title={Tree of thoughts: Deliberate problem solving with large language models},
  author={Yao, Shunyu and Yu, Dian and Zhao, Jeffrey and Shafran, Izhak and Griffiths, Tom and Cao, Yuan and Narasimhan, Karthik},
  journal={Advances in neural information processing systems},
  volume={36},
  pages={11809--11822},
  year={2023}
}

@inproceedings{lan2024stance,
  title={Stance detection with collaborative role-infused llm-based agents},
  author={Lan, Xiaochong and Gao, Chen and Jin, Depeng and Li, Yong},
  booktitle={Proceedings of the international AAAI conference on web and social media},
  volume={18},
  pages={891--903},
  year={2024}
}

@inproceedings{wu2024autogen,
  title={Autogen: Enabling next-gen LLM applications via multi-agent conversations},
  author={Wu, Qingyun and Bansal, Gagan and Zhang, Jieyu and Wu, Yiran and Li, Beibin and Zhu, Erkang and Jiang, Li and Zhang, Xiaoyun and Zhang, Shaokun and Liu, Jiale and others},
  booktitle={First conference on language modeling},
  year={2024}
}

@article{li2023camel,
  title={Camel: Communicative agents for" mind" exploration of large language model society},
  author={Li, Guohao and Hammoud, Hasan and Itani, Hani and Khizbullin, Dmitrii and Ghanem, Bernard},
  journal={Advances in neural information processing systems},
  volume={36},
  pages={51991--52008},
  year={2023}
}

@article{hong2023metagpt,
  title={MetaGPT: Meta Programming for A Multi-Agent Collaborative Framework},
  author={Hong, Sirui and Zhuge, Mingchen and Chen, Jiaqi and Zheng, Xiawu and Cheng, Yuheng and Zhang, Ceyao and Wang, Jinlin and Wang, Zili and Yau, Steven Ka Shing and Lin, Zijuan and Zhou, Liyang and Ran, Chenyu and Xiao, Lingfeng and Wu, Chenglin and Schmidhuber, Jürgen},
  journal={arXiv preprint arXiv:2308.00352},
  year={2023}
}

@inproceedings{du2024improving,
  title={Improving factuality and reasoning in language models through multiagent debate},
  author={Du, Yilun and Li, Shuang and Torralba, Antonio and Tenenbaum, Joshua B and Mordatch, Igor},
  booktitle={Forty-first international conference on machine learning},
  year={2024}
}

@inproceedings{wang2024deem,
  title={DEEM: Dynamic Experienced Expert Modeling for Stance Detection},
  author={Wang, Xiaolong and Wang, Yile and Cheng, Sijie and Li, Peng and Liu, Yang},
  booktitle={Proceedings of the 2024 Joint International Conference on Computational Linguistics, Language Resources and Evaluation (LREC-COLING 2024)},
  year={2024},
  pages={4530--4541}
}

@inproceedings{taranukhin2024stancereasoner,
  title={Stance Reasoner: Zero-Shot Stance Detection on Social Media with Explicit Reasoning},
  author={Taranukhin, Maksym and Shwartz, Vered and Milios, Evangelos},
  booktitle={Proceedings of LREC-COLING 2024},
  year={2024},
  pages={15257--15272}
}

@inproceedings{besta2024graph,
  title={Graph of thoughts: Solving elaborate problems with large language models},
  author={Besta, Maciej and Blach, Nils and Kubicek, Ales and Gerstenberger, Robert and Podstawski, Michal and Gianinazzi, Lukas and Gajda, Joanna and Lehmann, Tomasz and Niewiadomski, Hubert and Nyczyk, Piotr and others},
  booktitle={Proceedings of the AAAI conference on artificial intelligence},
  volume={38},
  number={16},
  pages={17682--17690},
  year={2024}
}

@inproceedings{garg2024stanceformer,
  title={Stanceformer: Target-aware transformer for stance detection},
  author={Garg, Krishna and Caragea, Cornelia},
  booktitle={Findings of the Association for Computational Linguistics: EMNLP 2024},
  pages={4969--4984},
  year={2024}
}

@inproceedings{yan2025collaborative,
  title={Collaborative stance detection via small-large language model consistency verification},
  author={Yan, Yu and Sun, Sheng and Tang, Zixiang and Liu, Teli and Liu, Min},
  booktitle={International Conference on Database Systems for Advanced Applications},
  pages={3--20},
  year={2025},
  organization={Springer}
}

@inproceedings{zhang2026msme,
  title={MSME: A Multi-Stage Multi-Expert Framework for Zero-Shot Stance Detection},
  author={Zhang, Yuanshuo and Li, Aohua and Chen, Bo and Sun, Jingbo and Zhao, Xiaobing},
  booktitle={Proceedings of the AAAI Conference on Artificial Intelligence},
  volume={40},
  number={41},
  pages={34879--34887},
  year={2026}
}

@inproceedings{riloff2013sarcasm,
  title={Sarcasm as Contrast between a Positive Sentiment and a Negative Situation},
  author={Riloff, Ellen and Qadir, Ashequl and Surve, Prafulla and De Silva, Lalindra and Gilbert, Nathan and Huang, Ruihong},
  booktitle={Proceedings of the Conference on Empirical Methods in Natural Language Processing (EMNLP)},
  pages={704--714},
  year={2013}
}

@inproceedings{bamman2015contextualized,
  title={Contextualized sarcasm detection on twitter},
  author={Bamman, David and Smith, Noah},
  booktitle={proceedings of the international AAAI conference on web and social media},
  volume={9},
  number={1},
  pages={574--577},
  year={2015}
}

@inproceedings{somasundaran2009stance,
  title={Recognizing Stances in Online Debates},
  author={Somasundaran, Swapna and Wiebe, Janyce},
  booktitle={Proceedings of the Joint Conference of the 47th Annual Meeting of the ACL and the 4th International Joint Conference on Natural Language Processing (ACL-IJCNLP)},
  year={2009},
  pages={226--234}
}

@inproceedings{shazeer2017moe,
  title={Outrageously Large Neural Networks: The Sparsely-Gated Mixture-of-Experts Layer},
  author={Shazeer, Noam and Mirhoseini, Azalia and Maziarz, Krzysztof and Davis, Andy and Le, Quoc V. and Hinton, Geoffrey and Dean, Jeff},
  booktitle={International Conference on Learning Representations (ICLR)},
  year={2017}
}

@inproceedings{lepikhin2021gshard,
  title={GShard: Scaling Giant Models with Conditional Computation and Automatic Sharding},
  author={Lepikhin, Dmitry and Lee, HyoukJoong and Xu, Yuanzhong and Chen, Dehao and Firat, Orhan and Huang, Yanping and Krikun, Maxim and Shazeer, Noam and Chen, Zhifeng},
  booktitle={International Conference on Learning Representations (ICLR)},
  year={2021}
}

@inproceedings{du2021glam,
  title={GLaM: Efficient Scaling of Language Models with Mixture-of-Experts},
  author={Du, Nan and Huang, Yanping and Dai, Andrew M. and Tong, Simon and Lepikhin, Dmitry and Xu, Yuanzhong and Krikun, Maxim and Zhou, Yanqi and Yu, Adams Wei and Firat, Orhan and Zoph, Barret and Fedus, Liam and Bosma, Maarten and Zhou, Zongwei and Wang, Tao and Wang, Yu Emma and Webster, Kellie and Pellat, Marie and Robinson, Kevin and Meier-Hellstern, Kathleen S. and Duke, Toju and Dixon, Lucas and Zhang, Kun and Le, Quoc V. and Wu, Yonghui and Chen, Z. and Cui, Claire},
  booktitle={International Conference on Machine Learning (ICML)},
  year={2021}
}

@inproceedings{zhou2023prompteng,
  title={Large Language Models are Human-Level Prompt Engineers},
  author={Zhou, Yongchao and Muresanu, Andrei Ioan and Han, Ziwen and Paster, Keiran and Pitis, Silviu and Chan, Harris and Ba, Jimmy},
  booktitle={International Conference on Learning Representations (ICLR)},
  year={2023}
}

@inproceedings{pryzant2023promptopt,
  title={Automatic Prompt Optimization with "Gradient Descent" and Beam Search},
  author={Pryzant, Reid and Iter, Dan and Li, Jerry and Lee, Yin Tat and Zhu, Chenguang and Zeng, Michael},
  booktitle={Proceedings of the Conference on Empirical Methods in Natural Language Processing (EMNLP)},
  year={2023}
}

@article{wei2022chain,
  title={Chain-of-thought prompting elicits reasoning in large language models},
  author={Wei, Jason and Wang, Xuezhi and Schuurmans, Dale and Bosma, Maarten and Xia, Fei and Chi, Ed and Le, Quoc V and Zhou, Denny and others},
  journal={Advances in neural information processing systems},
  volume={35},
  pages={24824--24837},
  year={2022}
}

\clearpage
\appendix

\setcounter{page}{1}
\setcounter{section}{0}
\setcounter{table}{0}
\setcounter{figure}{0}
\newcounter{offset}
  \setcounter{offset}{\value{figure}}
  \renewcommand{\thefigure}{S\the\numexpr\value{figure}-\value{offset}\relax}
  \setcounter{offset}{\value{table}}
  \renewcommand{\thetable}{T\the\numexpr\value{table}-\value{offset}\relax}

\section{Full Prompt Templates}
\label{app:prompts}
We provide the full prompt templates used for the Manager
(planning/decision) and Workers.

\subsection{Manager Prompt: Planning Stage}
\label{app:prompt_planning}

\begin{promptbox}{A1. Manager Prompt (Planning)}
\footnotesize

\textbf{You are the Manager in the planning stage.}

\vspace{4pt}
\textbf{Goal:} Decide the number of Workers and assign each a distinct analytical perspective.

\vspace{4pt}
\textbf{Input:}
\begin{itemize}
  \item Tweet text
  \item Stance target
  
\end{itemize}

\vspace{4pt}
\textbf{Instructions:}
\begin{enumerate}
  \item Judge instance complexity: \emph{simple}, \emph{medium}, or \emph{complex}.
        Treat as complex when sarcasm, indirect stance, mixed sentiment, multiple entities,
        or rhetorical framing is present.
  \item Select the number of Workers $N \in \{3,\dots,7\}$ (more for complex, fewer for simple).
  \item Write $N$ Worker prompts, each emphasizing a \emph{different} stance-relevant aspect
        (e.g., explicit cues, sarcasm/irony, emotional tone, indirect stance via related actors,
        policy framing, hashtags/slogans).
\end{enumerate}

\vspace{4pt}
\textbf{Output format (exact):}
\begin{itemize}
  \item \texttt{Number of Workers: N}
  \item \texttt{Worker 1 Prompt: ...}
  \item \texttt{Worker 2 Prompt: ...}
  \item \dots
\end{itemize}

\vspace{2pt}
\textbf{Constraint:} Do \textbf{not} output a stance label.
\end{promptbox}

\subsection{Manager Prompt: Decision Stage}
\label{app:prompt_decision}

\begin{promptbox}{A2. Manager Prompt (Decision)}
\footnotesize

\textbf{You are the Manager in the decision stage.}

\textbf{Goal:} Aggregate Worker reasonings and output a single final stance label.

\textbf{Input:}
\begin{itemize}[leftmargin=*]
    \item Tweet text
    \item Stance target
    \item Worker reasonings (each is 2--4 sentences; no final label)
\end{itemize}

\textbf{Instructions:}
\begin{enumerate}[leftmargin=*]
    \item Compare Worker reasonings and identify areas of agreement, disagreement, and the strongest evidence.
    
    \item Prioritize reasonings that cite concrete textual evidence from the tweet and directly relate it to the target.
    
    \item Discount reasonings that are unsupported, overly speculative, or rely only on general sentiment unrelated to the target.
    
    \item Resolve conflicts by selecting the label best supported by the combined evidence and target-specific stance interpretation.
    
    \item Select exactly one label from the valid set for the dataset:
    \begin{itemize}[leftmargin=*]
        \item \textbf{P-Stance:} {\it favor} / {\it against}
        \item \textbf{COVID-19:} {\it favor} / {\it against} / {\it none}
        \item \textbf{SemEval-2016:} {\it favor}  / {\it against} / {\it none}
    \end{itemize}
\end{enumerate}

\textbf{Output (exact):} Print only one valid stance label for the dataset.
\end{promptbox}

\subsection{Worker Prompt Template}
\label{app:prompt_worker}

\begin{promptbox}{A3. Worker Prompt (Template)}
\footnotesize

\textbf{You are a Worker.}

\vspace{4pt}
\textbf{Input:}
\begin{itemize}
  \item Tweet text
  \item Stance target
  
  \item Your assigned aspect (e.g., explicit cues, sarcasm/irony, emotional tone, indirect stance via related actors, framing, hashtags)
\end{itemize}

\vspace{4pt}
\textbf{Instructions:}
\begin{enumerate}
  \item Analyze stance toward the target focusing only on your assigned aspect.
  \item Write \textbf{2--4 sentences} of evidence-grounded reasoning.
\end{enumerate}

\vspace{2pt}
\textbf{Constraint:} Output reasoning only (no labels such as {\it favor}, {\it against}, {\it none}).
\end{promptbox}

\section{Dataset Examples}
\label{app:dataset_examples}

Table~\ref{tab:dataset_examples} presents three representative examples from the evaluated stance detection datasets together with their corresponding targets and labels.

\begin{table*}[t] \centering \small \caption{Sample instances from the evaluated stance detection datasets with their corresponding targets and labels.} \label{tab:dataset_examples} \setlength{\tabcolsep}{5pt} \renewcommand{\arraystretch}{1.08} \begin{tabular}{p{2.4cm}p{3.0cm}p{7.2cm}p{1.5cm}} \toprule \textbf{Dataset} & \textbf{Target} & \textbf{Tweet} & \textbf{Label} \\ \midrule SemEval-2016 & Climate Change & `Global Warming! Global Warming! Global Warming!  Oh wait, it's summer.` & {\it against}  \\ P-Stance & Bernie Sanders & `Bernie tell the people the truth u lie to get votes` & {\it against}  \\ COVID-19 Stance & Face Mask & `@IAmHighOnHeels \#Masks are important in the fight against \#Covid\_19 Come on\!` & {\it favor} \\ \bottomrule \end{tabular} \end{table*}

\section{Accuracy Results}
\label{sec:accuracy_results}

Table~\ref{tab:acc} summarizes accuracy across SemEval-2016, P-Stance, and COVID-19. The trends closely mirror Macro-F1, indicating consistent improvements across both class-balanced and overall metrics. Mistral achieves the highest average accuracy on SemEval-2016, while Gemini performs best on COVID-19.
Overall, gains are more pronounced on datasets requiring implicit or context-dependent interpretation (e.g., COVID-19), and smaller on more explicit datasets such as P-Stance, suggesting that reasoning-based aggregation is particularly beneficial for complex stance inference.

\begin{table*}[!t]
\centering
\begingroup
\fontsize{9.6}{12.9}\selectfont
\resizebox{\textwidth}{!}{%
\begin{tabular}{lcccccc|cccc|ccccc}
\toprule
\multirow{2}{*}{Model} &
\multicolumn{6}{c|}{SemEval16 (\%)} &
\multicolumn{4}{c|}{P-Stance (\%)} &
\multicolumn{5}{c}{COVID-19 (\%)} \\
\cmidrule(lr){2-7}\cmidrule(lr){8-11}\cmidrule(lr){12-16}
 & A & CC & FM & HC & LA & Avg $\pm$ std
 & DT & JB & BS & Avg $\pm$ std
 & FM & AF & SH & SC & Avg $\pm$ std \\
\midrule

Gemini
 & \textbf{91.15} & 94.78 & \textbf{83.82} & 90.32 & 73.62 & 86.74 $\pm$ 1.75
 & \textbf{90.95} & 85.82 & 76.01 & 84.26 $\pm$ 1.75
 & \textbf{87.00} & \textbf{81.50} & \textbf{86.50} & \textbf{88.50} & \textbf{85.88 $\pm$ 0.65} \\

Llama
 & 89.36 & \textbf{95.18} & 77.87 & 92.04 & 77.18 & 86.33 $\pm$ 0.78
 & 89.96 & \textbf{85.90} & \textbf{82.94} & \textbf{86.27 $\pm$ 0.41}
 & 86.00 & 79.00 & 80.50 & 83.67 & 82.29 $\pm$ 0.94 \\

Mistral
 & 88.02 & 93.28 & 80.91 & \textbf{93.55} & \textbf{82.13} & \textbf{87.58 $\pm$ 0.22}
 & 88.68 & 83.58 & 81.78 & 84.68 $\pm$ 0.22
 & 82.83 & 76.50 & 78.83 & 70.50 & 77.17 $\pm$ 0.71 \\
\bottomrule
\end{tabular}%
}
\endgroup
\vspace{-2mm}
\caption{
Accuracy (\%) on SemEval-2016 Task 6 \cite{mohammad2016semeval},
P-Stance \cite{li2021p}, and COVID-19 Stance \cite{glandt2021covidstance}
under the SMART-D framework.
Results are reported as mean $\pm$ standard deviation over three runs.
Best results are shown in \textbf{bold}.
}
\label{tab:acc}
\end{table*}

\section{Statistical Significance Analysis}
\label{app:statistical_significance}

We compare SMART-D against the corresponding single-model baselines on the
COVID-19 Stance dataset, where SMART-D shows the largest gains. For each
backbone, we compute Macro-F1 over three independent runs and evaluate the
difference using Welch's $t$-test. Because the number of runs is small
($n=3$), these tests should be interpreted as supporting evidence rather than
as a definitive estimate of significance.

Table~\ref{tab:significance} shows that SMART-D improves over the
corresponding single-model baseline for all three backbones. The gains are
largest for Llama and Mistral, and all improvements exceed
the observed run-to-run variance.

\begin{table}[!t]
\centering
\fontsize{8}{9.5}\selectfont
\setlength{\tabcolsep}{4pt}
\begin{tabular}{lcccc}
\toprule
\textbf{Backbone} & \textbf{Baseline} & \textbf{SMART-D} & $\Delta$ & \textbf{$p$-value} \\
\midrule
Gemini & 82.08 $\pm$ 0.05 & 86.07 $\pm$ 0.38 & +3.99 & 0.0026 \\
Llama          & 69.78 $\pm$ 0.29 & 81.31 $\pm$ 0.88 & +11.53 & 0.0008 \\
Mistral         & 67.79 $\pm$ 0.17 & 75.99 $\pm$ 0.33 & +8.20 & 0.00004 \\
\bottomrule
\end{tabular}
\caption{
Statistical significance of SMART-D improvements over corresponding
single-model baselines on the COVID-19 Stance dataset. Results are reported
as average Macro-F1  $\pm$ standard deviation over three runs. $\Delta$
denotes the absolute Macro-F1 improvement. Significance is evaluated using
Welch's $t$-test with three independent runs per method.
}
\label{tab:significance}
\end{table}


\section{Full Worker-Count Ablation}
\label{app:worker_count_full}

Table~\ref{tab:ablation_workers_all_full} reports the full fixed-$N$
Worker-count ablation across all datasets and backbones. Fixed
configurations use $N \in \{1,3,5,7,9,11\}$ Workers, while the dynamic
setting selects $N \in \{3,\ldots,7\}$ per instance. These results provide
the complete values summarized in Table~\ref{tab:worker_summary}.

\begin{table*}[t]
\centering
\small
\fontsize{6.8}{9.5}\selectfont
\setlength{\tabcolsep}{4pt}
\renewcommand{\arraystretch}{1.05}

\begin{tabular}{llcccccccc}
\toprule
\textbf{Dataset} & \textbf{Model} & $N{=}1$ & $N{=}3$ &$N{=}5$ & $N{=}7$ & $N{=}9$ & $N{=}11$ & Dyn. & $\bar{N}$ \\
\midrule

\multirow{3}{*}{COVID-19}
& Gemini  & 85.24$\pm$0.61 & 85.10$\pm$0.08 & 85.31$\pm$0.38 & 
85.64$\pm$0.51 & 85.34$\pm$0.23 & 85.56$\pm$0.86 & \textbf{86.07$\pm$0.38} & 3.60$\pm$0.00 \\
& Llama           & 79.50$\pm$0.38 & 80.76$\pm$0.71& 80.89$\pm$0.57 & 80.45$\pm$0.27 & 79.19$\pm$0.48 & \textbf{81.46$\pm$0.17} & 81.31$\pm$0.88 & 4.83$\pm$0.00\\
& Mistral          & 74.63$\pm$0.89 & 74.42$\pm$0.18 & 75.00$\pm$0.57 & \textbf{77.09$\pm$0.33} & 76.99$\pm$0.28 & 76.47$\pm$1.52 & 75.99$\pm$0.33 & 4.82$\pm$0.00 \\
\midrule

\multirow{3}{*}{P-Stance}
& Gemini  & 86.58$\pm$0.43 & 86.28$\pm$0.23 & 86.31$\pm$0.13 & 87.01$\pm$0.08 & 86.95$\pm$0.28 & \textbf{87.19$\pm$0.69} & 86.84$\pm$0.37 & 4.54$\pm$0.00 \\
& Llama           & 84.31$\pm$0.09 & 85.41$\pm$0.62 & 85.91$\pm$0.18& \textbf{86.27$\pm$0.18} & 86.22$\pm$0.33 & 86.15$\pm$0.01 & 85.96$\pm$0.24 & 5.00$\pm$0.00 \\
& Mistral          & 82.89$\pm$0.00 & 84.14$\pm$0.57 & 85.31$\pm$0.38& 85.00$\pm$0.38 & 85.61$\pm$0.37 & \textbf{85.71$\pm$0.12} & 85.48$\pm$0.15 & 5.13$\pm$0.01 \\
\midrule

\multirow{3}{*}{SemEval16}
& Gemini  & 82.55$\pm$0.24 & \textbf{82.95$\pm$0.64} & 82.97$\pm$0.57& 82.43$\pm$0.04 & 82.49$\pm$0.45 & 82.34$\pm$1.03 & 82.77$\pm$0.74 & 4.68$\pm$0.00 \\
& Llama           & 81.94$\pm$0.70 & 82.07$\pm$0.59 & 82.91$\pm$0.37& 82.89$\pm$0.35 & 84.31$\pm$0.91 & \textbf{84.92$\pm$0.21} & 82.90$\pm$0.26 & 4.82$\pm$0.01 \\
& Mistral          & 77.99$\pm$0.13 & 80.13$\pm$0.05 & 80.89$\pm$0.45& \textbf{81.96$\pm$0.06} & 81.90$\pm$0.02 & 81.67$\pm$1.25 & 80.76$\pm$0.45 & 4.00$\pm$0.00 \\
\bottomrule
\end{tabular}
\caption{
Worker allocation ablation across three stance benchmarks (Macro-F1, \%).
Fixed configurations use $N \in \{1,3,5,7,9,11\}$ Workers. Results are
reported as mean $\pm$ standard deviation over three runs. For the dynamic
setting, $\bar{N}$ denotes the average number of selected Workers. The
best result in each row is shown in \textbf{bold}.
}
\label{tab:ablation_workers_all_full}
\end{table*}

\section{Performance--Cost Trade-off}
\label{app:tradeoff}

\begin{figure*}[t]
\centering
\includegraphics[width=0.85\textwidth]{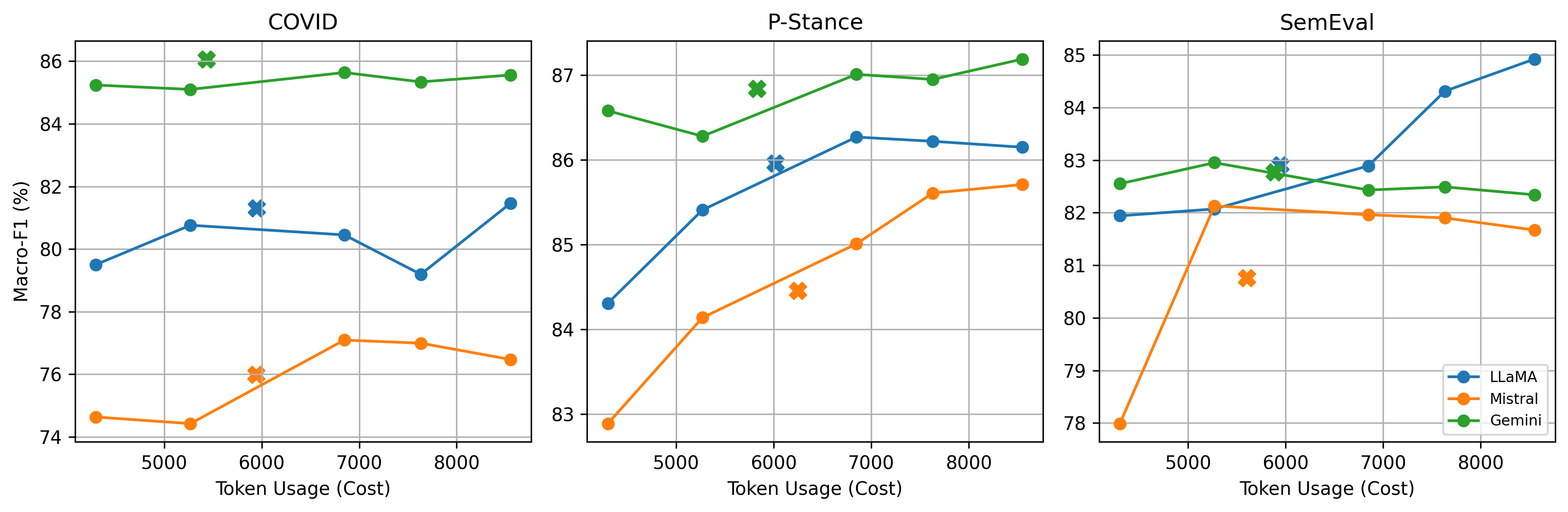}
\caption{
Performance--cost trade-off across three datasets. Lines represent fixed
Worker configurations, while star markers denote dynamic allocation. The
dynamic strategy achieves performance close to the best fixed
configurations while operating at lower computational cost.
}
\label{fig:tradeoff}
\end{figure*}

\begin{figure}[!t]
\centering
\includegraphics[width=0.8\linewidth]{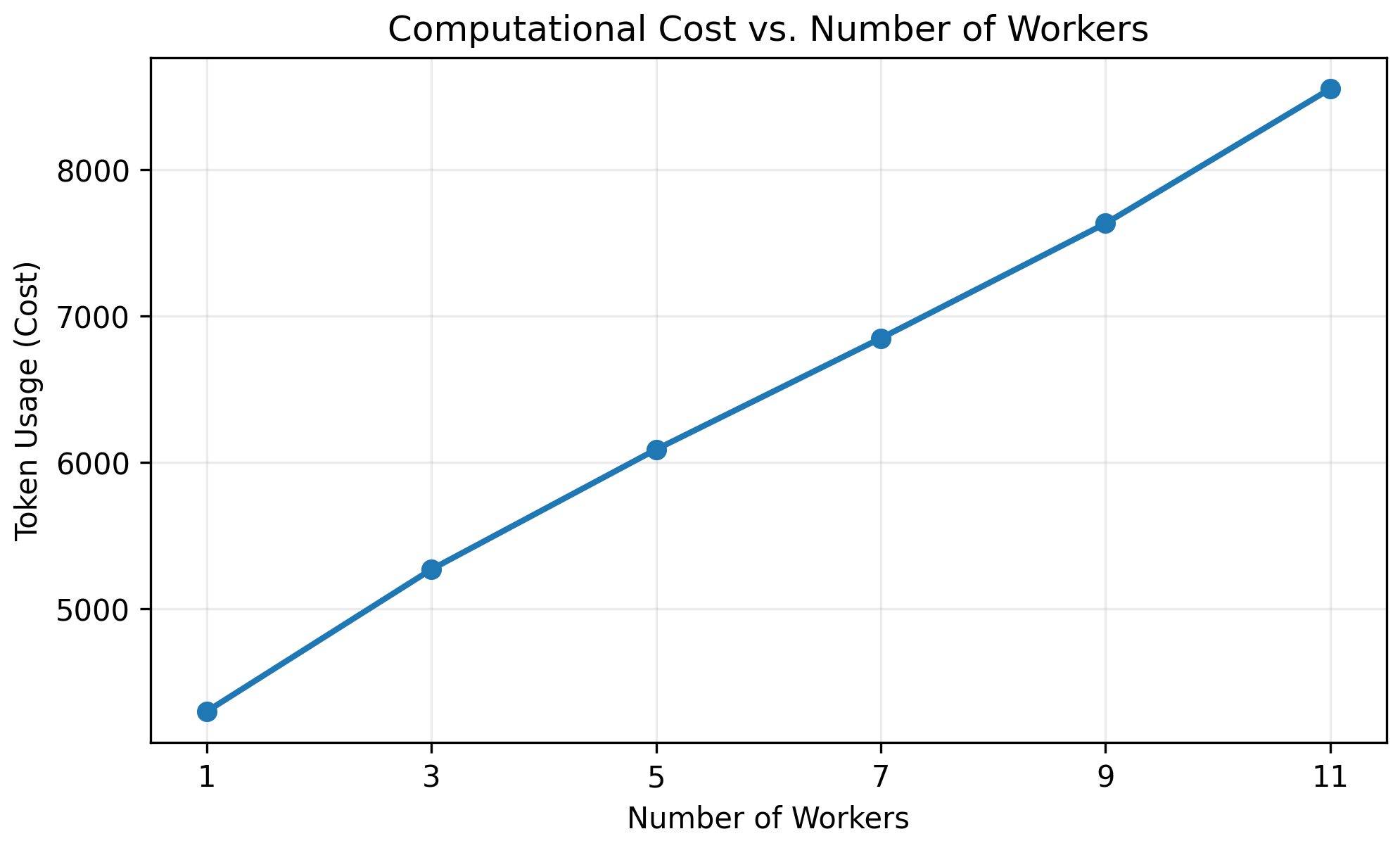}
\caption{
Token consumption grows approximately linearly with the number of Workers,
indicating a predictable computational overhead per Worker.
}
\label{fig:cost}
\end{figure}

Figure~\ref{fig:tradeoff} illustrates the performance--cost trade-off across three datasets.
Performance generally improves as the number of Workers ($N$) increases, but the gains tend to diminish beyond a moderate range, while computational cost continues to grow.  Figure~\ref{fig:cost} shows token usage as a function of the number of Workers. Token consumption grows approximately linearly with $N$, indicating a predictable computational overhead per additional Worker. As a result, increasing reasoning depth leads to higher inference cost even when performance improvements become limited. This supports the use of dynamic allocation, which aims to avoid unnecessary reasoning paths when additional Workers provide limited performance gains.

\section{Additional Dynamic Allocation Analysis}
\label{app:additional}

\subsection{Per-Target Worker Allocation}
\label{app:allocation_statistics}

This section provides detailed per-target distributions of dynamically
selected Worker counts, complementing the summary analysis in
Section~\ref{sec:worker_analysis}. The main text reports aggregate allocation
patterns; here, we show how the Manager's allocation decisions vary across
targets for the two representative settings used in Figure~\ref{fig:worker_distribution}.

\paragraph{COVID-19 Stance with Gemini.}
Table~\ref{tab:covid_allocation_details} shows the per-target allocation
distribution for COVID-19 Stance. Allocations are concentrated in the lower
range, primarily $N=3$ and $N=4$, with $N=5$ used selectively. This pattern is
consistent with the relatively low average allocation observed for this
setting.

\begin{table}[t]
\centering
\small
\setlength{\tabcolsep}{4pt}
\begin{tabular}{lrrrr}
\toprule
\textbf{Target} & \textbf{Inst.} & $N=3$ & $N=4$ & $N=5$ \\
\midrule
FM & 200 & 132 & 63  & 5  \\
SC   & 200 & 56  & 107 & 37 \\
AF      & 200 & 92  & 96  & 12 \\
SH & 200 & 94  & 88  & 18 \\
\bottomrule
\end{tabular}
\caption{
Per-target distribution of dynamically selected Worker counts for
COVID-19 Stance using Gemini. The four targets are: Wearing a Face Mask (FM), Anthony
Fauci (AF), Stay-at-Home Orders (SH), Keeping Schools Closed (SC).  Counts indicate the number of
instances assigned to each Worker count.
}
\label{tab:covid_allocation_details}
\end{table}

\paragraph{P-Stance with Mistral.}
Table~\ref{tab:pstance_allocation_details} shows the corresponding
distribution for P-Stance using Mistral. Compared with the COVID-19
setting, allocations shift toward larger Worker counts, especially $N=5$ and
$N=6$. This suggests that the Manager requests more reasoning perspectives
when using this backbone, even though P-Stance contains many more explicit
stance cues.

\begin{table}[t]
\centering
\small
\setlength{\tabcolsep}{4pt}
\begin{tabular}{lrrrrrr}
\toprule
\textbf{Target} & \textbf{Inst.} & $N=3$ & $N=4$ & $N=5$ & $N=6$ & $N=7$ \\
\midrule
JB      & 745 & 36 & 278 & 423 & 7   & 1  \\
DT   & 796 & 23 & 39  & 364 & 309 & 61 \\
BS & 635 & 6  & 13  & 333 & 270 & 13 \\
\bottomrule
\end{tabular}
\caption{
Per-target distribution of dynamically selected Worker counts for P-Stance
using Mistral. JB denotes {\it Joe Biden}, DT denotes {\it Donald Trump}, and BS denotes {\it Bernie Sanders}.  Counts indicate the number of instances assigned to each
Worker count.
}
\label{tab:pstance_allocation_details}
\end{table}

\paragraph{Summary.}
Across the two settings, Worker allocation varies by both dataset and
backbone. For COVID-19 Stance with Gemini, most instances are
assigned $N=3$ or $N=4$, indicating that the Manager often finds a small
number of reasoning perspectives sufficient. In contrast, for P-Stance with
Mistral, allocations shift toward $N=5$ and $N=6$, suggesting greater
model-dependent demand for additional reasoning perspectives.

Overall, these results support the interpretation that dynamic allocation is
not a simple proxy for dataset difficulty. Instead, the selected number of
Workers reflects an interaction between input characteristics and the
underlying model's reasoning behavior.
\subsection{Planning-Stage Examples}
\label{app:planning_examples}

To illustrate how the Manager allocates reasoning effort, we present three
planning-stage examples corresponding to low-, moderate-, and high-complexity
inputs. These examples show how the selected number of Workers reflects target
relevance, ambiguity, and the number of plausible interpretations.

\begin{planningexample}{Example 1: Low Complexity/Off-Topic Input ($N=3$)}

\small
\textbf{Tweet:} \emph{``New York city girl @fireboydml On \#trendmuzikblast
@Djitchyfingazz \#spinning:repeat\_one: LISTENLIVEON
\#OURAPPONPLAYSTORE AND \#WEBSITEONBIO''}

\vspace{3pt}
\textbf{Target:} Stay at Home Orders \\
\textbf{Gold label:} \emph{none} \\
\textbf{Baseline prediction:} \emph{none} \\
\textbf{SMART-D prediction:} \emph{none} \\
\textbf{Selected Workers:} 3

\vspace{4pt}
\textbf{Baseline reasoning.}
The tweet is a promotional post for a music broadcast and does not mention or
imply a stance toward Stay at Home Orders.

\vspace{4pt}
\textbf{Manager planning rationale.}
\begin{rationalebox}
\small
The tweet is primarily about music and promotion. There is no explicit mention
of, nor any obvious implication relating to, ``Stay at Home Orders.'' The
content does not express any stance toward the target, either directly or
indirectly. The main task here is simply to confirm that the tweet is
irrelevant to the target. Since there is no ambiguity, no implicit stance, and
no competing interpretations, three Workers are sufficient to confirm this
consensus and rule out any subtle interpretations.
\end{rationalebox}

\textbf{Interpretation.}
\begin{takeawaybox}
\small
The tweet is clearly unrelated to the target. The reasoning task is minimal:
confirm irrelevance and rule out a hidden target connection. Additional
Workers would likely be redundant.
\end{takeawaybox}

\end{planningexample}

\begin{planningexample}{Example 2: Moderate Complexity/Implicit but Recoverable Stance ($N=4$)}

\small
\textbf{Tweet:} {\small\emph{``@SpeakerPelosi @RepDanKildee \\@RepRichardNeal @RepDannyDavis Back to work!''}}

\vspace{3pt}
\textbf{Target:} Stay at Home Orders \\
\textbf{Gold label:} \emph{none} \\
\textbf{Baseline prediction:} \emph{against} \\
\textbf{SMART-D prediction:} \emph{none} \\
\textbf{Selected Workers:} 4

\vspace{4pt}
\textbf{Baseline reasoning.}
The phrase ``Back to work!'' directly advocates for ending lockdowns and
resuming normal activity, indicating opposition to stay-at-home orders.

\vspace{4pt}
\textbf{Manager planning rationale.}
\begin{rationalebox}
\small
The phrase ``Back to work!'' directed at several political figures strongly
suggests opposition to measures that prevent people from working, such as Stay
at Home Orders. However, the tweet does not explicitly mention the target
policy, and the stance must be inferred from context and intent. The brevity
of the tweet and the lack of direct reference to the policy introduce some
ambiguity. Additionally, the presence of multiple tagged accounts adds a
slight layer of contextual complexity, as the statement may be directed at
specific actions or broader policy decisions. While the stance is not deeply
ambiguous, it requires interpretation, and multiple reasoning perspectives are
needed to ensure the correct inference. More than four Workers would likely be
redundant, while fewer might miss subtle contextual cues.
\end{rationalebox}

\textbf{Interpretation.}
\begin{takeawaybox}
\small
The tweet contains an implicit cue, but the target connection is indirect.
Four Workers provide enough reasoning diversity to distinguish a general
political exhortation from a direct stance toward Stay at Home Orders,
leading to a target-aware \emph{none} prediction.
\end{takeawaybox}

\end{planningexample}

\begin{planningexample}{Example 3: High Complexity/Competing Interpretations ($N=5$)}

\small
\textbf{Tweet:}
\emph{``Better to be safe than sorry. \#OnlineClasses offer continued learning
in a safe environment. Let schools function and provide education. It's easy
to destroy with such senseless orders.''}

\vspace{3pt}
\textbf{Target:} School Closures \\
\textbf{Gold label:} \emph{favor} \\
\textbf{Baseline prediction:} \emph{favor} \\
\textbf{SMART-D prediction:} \emph{favor} \\
\textbf{Selected Workers:} 5

\vspace{4pt}
\textbf{Baseline reasoning.}
The tweet focuses on the safety benefits of online classes and continued
learning in a safe environment, which supports school closures. However, the
reference to ``senseless orders'' introduces ambiguity about whether the author
is criticizing closures, their implementation, or broader policy consequences.

\vspace{4pt}
\textbf{Manager planning rationale.}
\begin{rationalebox}
\small
While the author explicitly states a preference for ``OnlineClasses,'' which
could suggest support for restrictions or closures, the latter part introduces
a critical element---a condemnation of ``senseless orders.'' This implies a
negative stance toward the policies associated with those orders. However, the
interpretation is not straightforward. The emotionally charged phrasing
introduces ambiguity and raises questions about what exactly is being
criticized. It is unclear whether the criticism is directed at the closures
themselves, the implementation of online learning, or broader socio-economic
consequences of the policies. The tweet contains multiple interpretive layers
and potentially conflicting signals. Because of this ambiguity, several
plausible interpretations exist, and multiple reasoning perspectives are
required to explore them. Assigning five Workers ensures sufficient coverage
of these competing interpretations, while additional Workers would likely
yield diminishing returns.
\end{rationalebox}

\textbf{Interpretation.}
\begin{takeawaybox}
\small
The tweet contains both support for safe online learning and criticism of
policy orders. Even though both the baseline and SMART-D predict the correct
label, the Manager assigns more Workers because the input contains conflicting
signals and multiple plausible readings.
\end{takeawaybox}

\end{planningexample}

\section{Per-Target Ablation Results}
\label{app:per_target_ablation}

In this section, we provide detailed per-target Macro-F1 results for the ablation of reasoning components on the COVID-19 stance dataset.

Table~\ref{tab:per_target_ablation} reports results across four targets:
Face Masks (FM), Anthony Fauci (AF), Stay-at-Home (SH), and School Closures (SC).
We include all configurations from the main text, along with the Self-Consistency baseline for completeness.

\begin{table*}[!t]
\centering
\small
\fontsize{7.8}{8.9}\selectfont

\begin{tabular}{l l c c c c c}
\toprule
Method & Model & FM & AF & SH & SC & Avg $\pm$ std \\
\midrule

\multirow{3}{*}{WV + MR}
 & Gemini & 85.10 & 78.39 & 81.37 & 68.91 & 78.45 $\pm$ 0.52 \\
 & Llama          & 68.24 & 62.32 & 69.49 & 30.30 & 57.59 $\pm$ 1.18 \\
 & Mistral        & 62.78 & 73.67 & 69.99 & 33.67 & 60.02 $\pm$ 1.71 \\
\midrule

\multirow{3}{*}{WV}
 & Gemini & 83.51 & 78.71 & 82.64 & 70.95 & 78.95 $\pm$ 1.41 \\
 & Llama          & 74.82 & 70.23 & 68.43 & 51.26 & 66.18 $\pm$ 0.78 \\
 & Mistral        & 65.90 & 73.20 & 68.98 & 40.72 & 62.20 $\pm$ 1.27 \\
\midrule

\multirow{3}{*}{WV + WR}
 & Gemini & 88.71 & 82.27 & 83.63 & 86.47 & 85.27 $\pm$ 0.62 \\
 & Llama          & 79.71 & 78.73 & 77.15 & 56.38 & 72.99 $\pm$ 0.72 \\
 & Mistral        & 78.70 & 74.47 & 76.23 & 55.29 & 71.17 $\pm$ 0.52 \\
\midrule

\multirow{3}{*}{WV + WR + MR}
 & Gemini & 87.60 & 83.65 & 81.79 & 86.96 & 85.00 $\pm$ 0.18 \\
 & Llama          & 83.19 & 77.87 & 77.65 & 61.71 & 75.10 $\pm$ 0.28 \\
 & Mistral        & 78.79 & 76.70 & 77.56 & 56.36 & 72.35 $\pm$ 0.31 \\
\midrule

\multirow{3}{*}{Self-Consistency}
 & Gemini & 88.47 & 79.75 & 77.42 & 83.79 & 82.36 $\pm$ 0.18 \\
 & Llama          & 84.70 & 77.51 & 82.09 & 38.15 & 70.61 $\pm$ 0.28 \\
 & Mistral        & 80.25 & 75.85 & 76.07 & 42.51 & 68.67 $\pm$ 0.33 \\
\midrule

\multirow{3}{*}{SMART-D (ours)}
 & Gemini & \textbf{88.37} & \textbf{82.52} & \textbf{86.03} & \textbf{87.35} & \textbf{86.07 $\pm$ 0.38} \\
 & Llama          & 84.60 & 79.49 & 78.68 & 82.47 & 81.31 $\pm$ 0.88 \\
 & Mistral        & 81.89 & 76.56 & 75.03 & 70.49 & 75.99 $\pm$ 0.33 \\
\bottomrule

\end{tabular}
\caption{
Per-target Macro-F1 (\%) results on the COVID-19 dataset.
WV: Worker Voting, WR: Worker Reasoning, MR: Manager Reasoning.
Average $\pm$ standard deviation is reported over three runs.
}
\label{tab:per_target_ablation}

\end{table*}

\section{Cross-Model Worker Selection Patterns}
\label{app:worker_model_distribution}

This section analyzes how the Manager composes heterogeneous Worker teams in
the mixed-model setting on the COVID-19 Stance dataset. The Worker pool
contains the three backbones used throughout the paper:
Gemini, Llama, and Mistral. This analysis
provides additional insight into when cross-model diversity is selected and
how selection patterns vary across Manager backbones and stance targets.

\begin{figure*}[t]
\centering
\includegraphics[width=\linewidth]{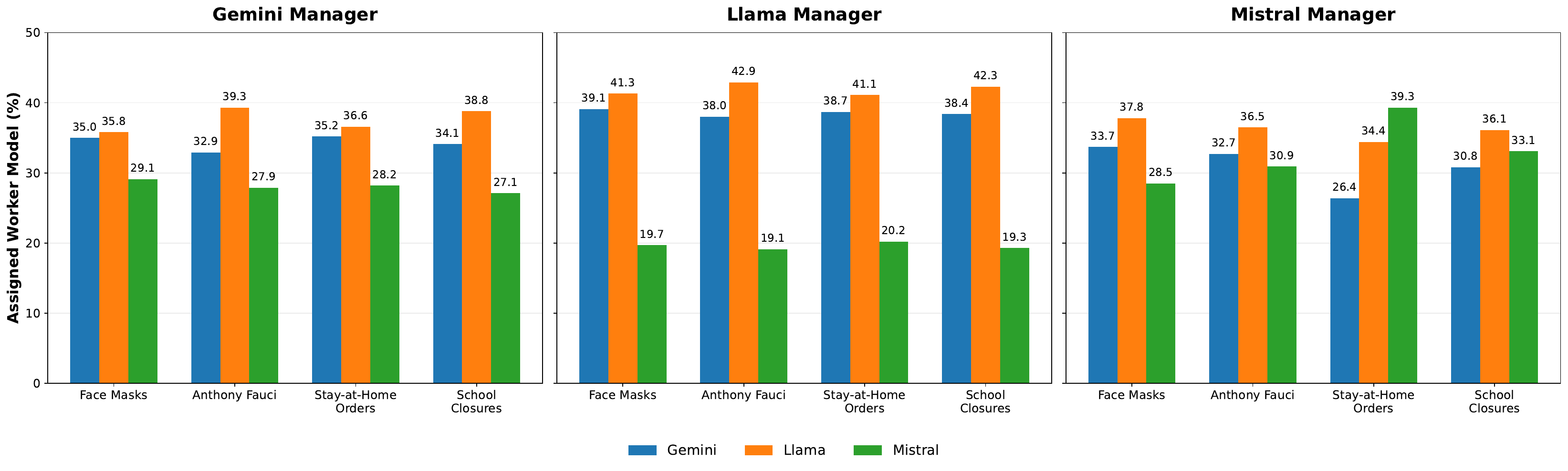}
\caption{
Distribution of selected Worker models in the mixed-model setting on
COVID-19 Stance. Panels correspond to Manager backbones:
Gemini, Llama, and Mistral. Bars show the
proportion of selected Workers from each backbone across four targets:
Face Masks (FM), Anthony Fauci (AF), Stay-at-Home Orders (SH), and School
Closures (SC).
}
\label{fig:worker_model_distribution}
\end{figure*}

Figure~\ref{fig:worker_model_distribution} shows that Worker selection is not
uniform across models. Instead, the selected Worker composition varies with
both the Manager backbone and the stance target. Stronger Managers, such as
Gemini and Llama, tend to rely more heavily on
stronger or similarly capable Workers, while still incorporating model
diversity in some target settings.

The selection behavior is also target-dependent. Targets such as
\textit{Stay-at-Home Orders} and \textit{School Closures} exhibit more varied
Worker compositions, suggesting that the Manager draws on complementary
reasoning signals when the target involves more implicit, policy-dependent,
or rhetorically complex stance expressions.

For Mistral, the mixed-model setting more frequently incorporates
stronger Worker models. This pattern is consistent with the main-text result
showing that Mistral benefits most from heterogeneous Workers. In contrast,
the stronger Manager backbones perform better in the homogeneous SMART-D
setting, suggesting that consistency in reasoning style can be more important
than model diversity when the Manager is already strong.

Overall, these results help explain why model diversity alone does not
guarantee better reasoning-level aggregation. Cross-model Worker selection can
provide useful complementary reasoning signals, particularly for weaker
backbones, but heterogeneous reasoning styles may also introduce inconsistency
that makes aggregation less stable. This behavior is related to
Mixture-of-Experts-style routing, but here the selection occurs at inference
time through Manager prompting rather than through a trained router or
additional supervision.

\section{Additional Qualitative Case Studies}
\label{app:case_studies}

This appendix provides the full qualitative examples referenced in
Section~\ref{sec:case_study}. For each case, we report the tweet, target,
gold label, baseline prediction, SMART-D prediction, selected number of
Workers, Worker prompts and reasonings, and the final Manager decision.

\begin{qualcase}{Case 1: Rhetorical Challenge and Implicit Support}
\small
\textbf{Tweet:}
\begin{quote}
\small\emph{``Governor @GovRonDeSantis, why aren't your offices open? If you
want all schools open in a month, open your offices today. \#COVID19''}
\end{quote}

\textbf{Target:} School Closures \\
\textbf{Gold label:} \emph{favor} \\
\textbf{Baseline prediction:} \emph{against} \\
\textbf{SMART-D prediction:} \emph{favor} \\
\textbf{Selected Workers:} 5

\vspace{4pt}

\begin{workerbox}{Worker 1: Direct Policy Stance}
\textbf{Prompt.}
Analyze the stance toward the policy of ``School Closures.'' Specifically,
evaluate whether the author's demand for the Governor to open his own offices
is a sincere endorsement of reopening or a rhetorical device used to highlight
potential risks. Focus on how the text conveys its position, considering
implicit cues such as framing, tone, or indirect references. Write a concise
2--4 sentence reasoning; do not output a final label.

\vspace{3pt}
\textbf{Reasoning.}
The author uses the demand for the Governor to open his offices as a rhetorical
challenge to highlight a perceived double standard rather than as a sincere
endorsement of reopening schools. By pointing out that the Governor's own
workspace remains closed, the author suggests that if it is not safe for
officials to return to the office, it is certainly not safe to mandate
in-person schooling. Consequently, the author is expressing opposition to the
policy of reopening schools during the pandemic.
\end{workerbox}

\begin{workerbox}{Worker 2: Rhetorical Strategy and Sarcasm}
\textbf{Prompt.}
Examine the use of sarcasm and rhetorical comparison in this tweet. Analyze
how the author frames the argument, particularly through indirect or ironic
expressions. Focus on identifying rhetorical cues and how they shape the
interpretation of the message, without assigning any explicit stance label.

\vspace{3pt}
\textbf{Reasoning.}
The author uses a logical comparison to highlight a perceived double standard,
suggesting that if schools are safe enough for a mandatory reopening, the
Governor's own administrative offices should already be operational. This
sarcastic challenge criticizes the reopening plan as hypocritical and
potentially unsafe by pointing out that the leader is not subject to the same
risks he is imposing on students and teachers. Rather than genuinely
advocating for more openings, the tweet uses the Governor's closed offices as
evidence that the push for open schools is inconsistent with current safety
protocols.
\end{workerbox}

\begin{workerbox}{Worker 3: Safety and Risk Assessment}
\textbf{Prompt.}
Analyze how the author frames COVID-19 safety concerns in the context of
schools. Does the author's challenge to the Governor suggest a comparison
between the safety of official workplaces and public schooling environments?
Explain how this perceived risk is expressed in the argument and how it shapes
the interpretation of the message. Write a concise 2--4 sentence reasoning;
do not output a final label.

\vspace{3pt}
\textbf{Reasoning.}
The author suggests that the Governor is applying a double standard by
demanding schools reopen while his own offices remain closed due to safety
concerns. This challenge implies that if the environment is not safe enough
for state officials, it is inherently too dangerous for students and teachers
to return to classrooms. Consequently, the author emphasizes safety
considerations, framing school closures as a precaution until public
environments are demonstrated to be sufficiently safe.
\end{workerbox}

\begin{workerbox}{Worker 4: Target Specificity and Relevance}
\textbf{Prompt.}
Distinguish between the author's criticism of the politician, Governor
DeSantis, and the specific policy context of ``School Closures.'' While the
tweet targets the Governor, focus on how this criticism relates to the broader
discussion of school reopening decisions. Analyze how the argument connects
these two elements and what implications are suggested regarding the policy
context. Write a concise 2--4 sentence reasoning; do not output a final label.

\vspace{3pt}
\textbf{Reasoning.}
The author highlights a potential inconsistency by questioning why the
Governor advocates reopening schools while his own offices remain closed.
This comparison connects criticism of the Governor to broader concerns about
safety and decision-making. The argument emphasizes a perceived mismatch
between policy expectations and actual practices, suggesting caution in
interpreting reopening decisions.
\end{workerbox}

\begin{workerbox}{Worker 5: Advocacy vs. Reporting}
\textbf{Prompt.}
Determine whether this tweet functions as a neutral description of a situation
or as an attempt to influence interpretation through a call to action. Analyze
the reference to the Governor's offices and how it is used within the
argument. Focus on whether the tweet frames the issue in a descriptive or
persuasive manner, and how this affects the interpretation of the message.
Write a concise 2--4 sentence reasoning; do not output a final label.

\vspace{3pt}
\textbf{Reasoning.}
The author uses a direct challenge to the Governor to highlight a perceived
inconsistency between school reopening plans and the status of government
workplaces. By calling for the Governor to ``open your offices today,'' the
tweet moves beyond neutral description and introduces a persuasive element.
This call to action functions as a rhetorical device to question the coherence
and safety of the reopening policy.
\end{workerbox}

\textbf{Manager decision.}
\begin{managerbox}
\small
Across the five Worker outputs, the dominant interpretation is that the tweet
does not literally call for broader reopening. Instead, it uses rhetorical
comparison to criticize reopening schools while government offices remain
closed. The Manager therefore predicts \emph{favor}, consistent with the gold
label.
\end{managerbox}

\end{qualcase}

\begin{qualcase}{Case 2: Distinguishing stance from related complaints}

\small
\textbf{Tweet:}
\begin{quote}
\small\emph{``What if there was job loss or pay cut due to recession? Would
government still give decision to not pay fees to schools? Are schools run by
robots?? Teachers also need to get paid right?''}
\end{quote}

\textbf{Target:} School Closures \\
\textbf{Gold label:} \emph{none} \\
\textbf{Baseline prediction:} \emph{against} \\
\textbf{SMART-D prediction:} \emph{none} \\
\textbf{Selected Workers:} 3

\vspace{4pt}

\begin{workerbox}{Worker 1: Policy Target vs. General School Issues}
\textbf{Prompt.}
Analyze whether the tweet focuses on the specific policy context of ``School
Closures'' or discusses broader issues related to education. Examine how
references such as ``not paying fees'' and ``teachers getting paid'' are used
in the argument, and whether they relate to operational, financial, or
policy-level considerations. Focus on how the content frames the topic and
what aspects of the issue are emphasized. Do not output a final label; provide
reasoning only.

\vspace{3pt}
\textbf{Reasoning.}
The author focuses on the economic sustainability of educational institutions,
questioning the fairness of fee waivers and the necessity of maintaining
teacher salaries during financial downturns. While the mention of not paying
fees often arises during periods of remote learning, the tweet centers on
financial and labor disputes rather than the safety or public health necessity
of closing school buildings. The reasoning is rooted in the economic impact on
staff and institutional operations rather than a specific policy stance on the
physical opening or closing of schools.
\end{workerbox}

\begin{workerbox}{Worker 2: Author Intent and Advocacy}
\textbf{Prompt.}
Examine the author's primary concern in this tweet. Analyze how references
such as ``schools run by robots'' and ``teachers needing to get paid'' are
used in the argument. Focus on whether the tweet emphasizes operational,
financial, or educational aspects, and how these elements shape the
interpretation of the message. Do not output a final label; provide reasoning
only.

\vspace{3pt}
\textbf{Reasoning.}
The author focuses on the financial viability of educational institutions,
emphasizing that teachers require salaries and schools have operational costs
that must be met even during economic downturns. While the tweet expresses
concern over tuition payment mandates during a recession, it primarily defends
the necessity of school funding rather than emphasizing a particular
instructional setting. The rhetoric regarding ``schools run by robots''
highlights concerns about staff livelihoods, without explicitly prioritizing a
specific operational mode.
\end{workerbox}

\begin{workerbox}{Worker 3: Contextual Relevance and Neutrality}
\textbf{Prompt.}
Analyze whether the tweet discusses issues directly related to the policy of
``School Closures'' or focuses on broader aspects such as funding, salaries,
or institutional concerns. Examine how the author's frustration with
government decisions on fees is framed, and whether it reflects policy-level
considerations or a separate economic concern. Focus on how the content
relates to the target context and what aspects of the issue are emphasized.
Do not output a final label; provide reasoning only.

\vspace{3pt}
\textbf{Reasoning.}
The tweet focuses on the economic implications of school fees and the
necessity of paying teacher salaries during a recession, rather than
emphasizing the operational status of schools. The author expresses
frustration with government mandates regarding tuition payments, framing the
issue as a matter of labor rights and institutional sustainability. Because
the discussion centers on financial and administrative concerns, it reflects
broader educational issues rather than directly addressing the policy context
of school closures.
\end{workerbox}

\textbf{Manager decision.}
\begin{managerbox}
\small
The Worker outputs consistently conclude that the tweet concerns school fees,
teacher pay, and financial logistics rather than a direct stance on reopening
or continued closure. The Manager therefore predicts \emph{none}, matching
the gold label.
\end{managerbox}

\end{qualcase}

\subsection{Example Execution Trace}
\label{app:trace}

To illustrate the full reasoning workflow, we provide a simplified execution
trace for Case A.

\begin{managerbox}
\small
\textbf{Manager planning output.}
\begin{itemize}
    \item Number of Workers: 5
    \item Worker 1: direct policy stance
    \item Worker 2: rhetorical strategy and sarcasm
    \item Worker 3: safety and risk assessment
    \item Worker 4: target specificity and relevance
    \item Worker 5: advocacy vs.\ reporting
\end{itemize}

\textbf{Worker outputs.}
Each Worker generates a short reasoning paragraph from its assigned
perspective. Some Workers focus on rhetorical inconsistency, while others
emphasize safety concerns or distinguish criticism of the Governor from stance
toward school closures.

\textbf{Manager decision.}
After aggregating the Worker reasonings, the Manager identifies the dominant
interpretation that the tweet criticizes reopening schools by highlighting a
double standard between government offices and schools. The final prediction
is therefore \emph{favor} for the target \emph{School Closures}.
\end{managerbox}

\end{document}